\documentclass[10pt,twocolumn,letterpaper]{article}

\usepackage{cvpr}              

\usepackage{graphicx}
\usepackage{amsmath}
\usepackage{amssymb}
\usepackage{booktabs}
\usepackage{bbding}
\usepackage{multirow}

\usepackage[dvipsnames,table]{xcolor}

\definecolor{cvprblue}{rgb}{0.21,0.49,0.74}
\usepackage[pagebackref,breaklinks,colorlinks,citecolor=cvprblue]{hyperref}

\def\paperID{1659} 
\def\confName{CVPR}
\def\confYear{2024}

\title{Towards Generalizable Multi-Object Tracking }

\author{Zheng Qin$^1$ ~Le Wang$^{1 *}$ ~Sanping Zhou$^1$ ~Panpan Fu$^2$ ~Gang Hua$^3$~Wei Tang$^4$ \\
	$^{1}$National Key Laboratory of Human-Machine Hybrid Augmented Intelligence, \\ National Engineering Research Center for Visual Information and Applications,  \\Institute of Artificial Intelligence and Robotics, Xi'an Jiaotong University\\
	$^{2}$School of Software Engineering, Xi'an Jiaotong University\\
	$^{3}$Wormpex AI Research ~
	$^{4}$University of Illinois at Chicago\\
}

\begin{document}
\maketitle
\begin{abstract}
	Multi-Object Tracking (MOT) encompasses various tracking scenarios, each characterized by unique traits. Effective trackers should demonstrate a high degree of generalizability across diverse scenarios. However, existing trackers struggle to accommodate all aspects or necessitate hypothesis and experimentation to customize the association information~(motion and/or appearance) for a given scenario, leading to narrowly tailored solutions with limited generalizability. In this paper, we investigate the factors that influence trackers' generalization to different scenarios and concretize them into a set of tracking scenario attributes to guide the design of more generalizable trackers. Furthermore, we propose a “point-wise to instance-wise relation” framework for MOT, i.e., GeneralTrack, which can generalize across diverse scenarios while eliminating the need to balance motion and appearance. Thanks to its superior generalizability, our proposed GeneralTrack achieves state-of-the-art performance on multiple benchmarks and demonstrates the potential for domain generalization. https://github.com/qinzheng2000/GeneralTrack.git
\end{abstract}
\footnotetext{$^*$Corresponding author.}    
\section{Introduction}
\label{sec:intro}

Multi-Object Tracking~(MOT) aims to locate targets and recognize their identities from a streaming video. It is an essential task for many applications such as autonomous driving~\cite{driving}, robotics~\cite{Robot}, and visual surveillance~\cite{surveil}. Despite great progress in the past few years, the MOT task remains challenging when the trackers are generalized to diverse application scenarios.

Prior MOT methods mostly follow the tracking-by-detection~(TbD)~\cite{tracking-by-detection,bytetrack,botsort} or tracking-by-regression~(TbR) \cite{centertrack, tracktor} paradigm. TbD methods detect objects in each frame and then associates objects across frames. TbR methods also conduct frame-wise object detection, but replaces the data association with a continuous regression of each tracklet to its new position.  
With rapid advances in object detection, TbD has become the dominant paradigm in the field.

\begin{figure*}[t]
	\centering
	\includegraphics[width=1\linewidth]{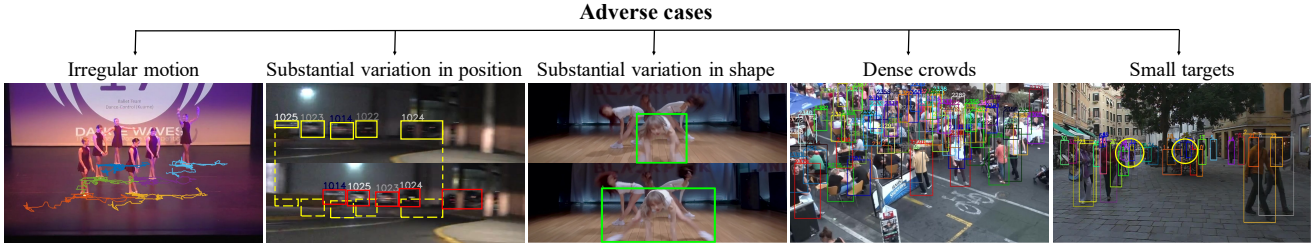}
	\caption{{\bf Adverse cases for some attributes in tracking scenarios.}. The line in 'irregular motion' is the trajectory of each target.}
	\label{fig:supp}
	
\end{figure*}

Real-world application scenarios are diverse and characterized by a wide spectrum of different attributes, such as varying motion complexities, target densities, and frame rate, as shown in~\Cref{fig:supp}.  Unfortunately, current MOT methods heavily depend on extensive prior knowledge or intricate engineering efforts to excel in specific scenarios, but they struggle to generalize effectively to different situations. This limitation significantly restricts their utility in real-world applications. 

For TbD, motion-dominated methods~\cite{sort,motiontrack,probabilistic} are brittle when encountering irregular motion and substantial variation in target shape or position; appearance-dominated methods~\cite{deepsort,strongsort,focus} are prone to failure when facing occlusion, for example, caused by dense crowds or shelters, light change, and small targets. To overcome these difficulties, some TbD methods require a \textit{manual} adjustment of which information to rely on more in a specific scenario. For example, ByteTrack~\cite{bytetrack} constructs an affinity matrix based on motion in MOT17 and MOT20, and based on appearance in BDD100K. Other works~\cite{seidenschwarz2023simple, strongsort, deepsort, botsort} directly balance the two affinity matrices with a weighting factor and adjust it for different scenarios. The limitation in generalizability similarly applies to TbR. For example, Tracktor~\cite{tracktor} cannot handle videos with low frame rates and targets with large shape or position variation;
Centertrack~\cite{centertrack} uses the center point to represent each target, which would become overwhelming in crowded scenarios.
Therefore, there is an urgent need to develop trackers that effectively generalize to different scenarios.

In this paper, we first conduct an in-depth analysis of the tracking scenarios to gain insight into why a particular tracker's performance varies significantly in different scenarios. Referring to the performance of previous trackers on different datasets, we parsed out the tracking scenario attributes as follows: motion complexity, variation amplitude, target density, small target, and frame rate. We analyze the datasets~\cite{mot16,mot20,bdd100k,dancetrack,sportsmot} commonly used in MOT based on these attributes, and find that these attributes vary drastically across different datasets. 
Both motion-dominated and appearance-dominated methods have their respective attributes which they do not excel at.

Based on the above analysis, we propose a “point-wise to instance-wise relation” framework for MOT, \textit{i.e.}, GeneralTrack, which can generalize to diverse scenarios without manually balancing motion and appearance information. Specifically, instead of directly constructing relations between tracklets and detections at the instance level, we capture the point-wise relations and then translate them into instance-wise associations. The fine-grained features together with the fine-to-coarse translation can cope with dense targets and small targets. In contrast to searching in fixed local areas by previous motion-dominated methods, we contrust multi-scale point-region relation which implicitly contains a motion template guided by vision and geometry that does not flinch at irregular motion. The flexible scale of motion templates can be effectively adapted to various frame rates as well as the amplitude of position variation. Finally, we design a hierarchical relation aggregation paradigm to associate the tracklets and detections according to the point-part-instance hierarchy. The targets evolve from rigid bodies to flexible bodies and are suitable for scenarios with dramatic shape variation.

Extensive experiments on multiple benchmark datasets show that our method achieves the state-of-the-art, demonstrating the superiority of generalizability over diverse scenarios. In particular, our GeneralTrack ranks 1st place on the BDD100K leaderboard~(57.87 mTETA). In addition, we experimentally find that GeneralTrack has great potential for domain generalization with unseen data distributions~(cross-dataset, cross-class). The main contribution of this work can be summarized as follows:
\begin{itemize}
	\item We analyze the factors that hinder the generalizability of existing trackers and concretize them into tracking scenario attributes that can guide the design of trackers.
    \item We propose a ``point-wise to instance-wise relation" framework for MOT. It first constructs point-wise relations through the multi-scale 4D correlation volume
    and then aggregates them into instance-wise associations through a novel  ``point-part-instance" hierarchy. Our new framework can address several fundamental challenges in MOT. Concretely, the point-wise correlation modeling deals with damage to instance-level representations by dense and small targets; the construction of multi-scale point-region relations handles severe motion complexity, and different position variations and frame rates; the hierarchical aggregation copes with shape variations.
    \item Extensive evaluation of the GeneralTrack shows that it achieves the state-of-the-art performance on multiple MOT datasets. In addition, GeneralTrack experimentally demonstrates strong domain generalization capabilities.
\end{itemize}

\section{Related Work}
\label{sec:Related Work}
\noindent{\bf Tracking-by-Detection.}
The dominant paradigm in the field of MOT has long been tracking-by-detection~\cite{bytetrack, sort, strongsort, ocsort, fft, he2021learnable, quasi, permatrack, li2023single}. The core of TbD is to construct inter-frame relation~(affinity matrix) between tracklets and detections, and then perform matching with the Hungarian Algorithm~\cite{hungarian}. The affinity matrix for matching is often driven by motion information~\cite{motiontrack, probabilistic,  Motion-aware, Detecting-invisible-people} or appearance information~\cite{deepsort, yu2022towards, kim2021discriminative, wang2021multiple, pang2021quasi}. As discussed in Sec. \ref{sec:intro}, both motion and appearance dominated methods have their respective scenarios in which they do not excel.

To address these issues, some methods work towards a better balance between motion and appearance~\cite{bytetrack, seidenschwarz2023simple, strongsort, deepsort, botsort}; some others, \textit{i.e.}, TrackFlow~\cite{mancusi2023trackflow}, handle these by building an probabilistic formulation but requires virtual datasets for training.
In contrast, we propose a new approach that achieves generalization and avoids balance between motion and appearance.

\noindent{\bf Dense Flow and Correspondences.}
Identifying correspondences between an image pair is a fundamental computer vision problem, encompassing optical flow and geometric correspondences. 
FlowNet~\cite{dosovitskiy2015flownet} is the first end-to-end method for optical flow estimation. Then a series of works~\cite{sun2018pwc, sun2019models, hui2018liteflownet, hui2020lightweight,ranjan2017optical,dong2023sparse} employ coarse-to-fine and iterative estimation methodology. To deal with missing small fast-motion objects in the coarse stage, RAFT~\cite{raft} performs optical flow estimation in a coarse-and-fine and recurrent manner. Geometric correspondences~\cite{lowe2004distinctive,revaud2019r2d2,sarlin2020superglue} refer to correspondences between images captured from different views. MatchFlow~\cite{dong2023rethinking} takes geometric correspondences as the prefixed task for optical flow.

Among these explorations, the 4D Correlation Volume is often used to capture the visual similarity of pixel pairs and as a core component supporting dense flow and correspondence estimation. In MOT, the visual relation between frames is significant. Inspired by these works, we address the tracking task from the perspective of constructing the relation from pixel to instance with low-level vision.

\section{Methodology}
\label{sec:Methodology}

\subsection{Analysis of MOT Scenarios}

\begin{figure}[t]
	\centering
	\includegraphics[width=0.9\linewidth]{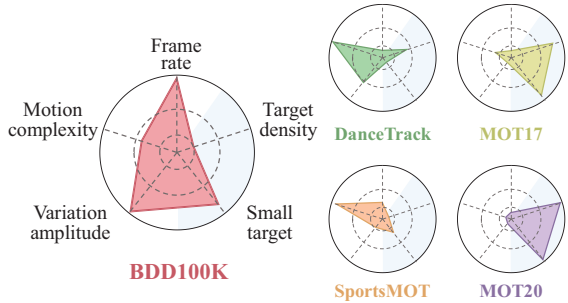}
	\caption{\textbf{Tracking scenario attribute maps}. Appearance performs poorly in the scenario with a large percentage in the blue area, as well as motion in the white area.}
	\label{fig:intro}
	
\end{figure}

There are countless real-world application scenarios. Current MOT methods heavily depend on extensive prior knowledge or intricate engineering efforts to excel in specific scenarios, but they hardly generalize to different situations. To gain insight into this phenomenon, we analyze the failure cases of previous trackers in different datasets and identify the following attributes that have substantial influence on a tracker's performance:

\begin{itemize}
	\item \noindent {\bf Motion Complexity}
 reflects the irregularity and unpredictability of target motion within the scenario. The more irregular and unpredictable the motion, the greater its complexity.
	\item \noindent {\bf Variation Amplitude} reflects the target's variability, encompassing both shape and position variations.
    \item \noindent {\bf Target Density} reflects the density of the crowds in the scenario, implicitly reflecting the degree of occlusion within the crowds.
    \item \noindent {\bf Small Target} represents the average amount of small targets in the scenario.
    \item \noindent {\bf Frame Rate} is
the number of frames captured in one second of the input video stream.
\end{itemize}
\noindent We conduct thorough measurements of these attributes on five datasets~\cite{mot16,mot20,bdd100k,dancetrack,sportsmot} and form the tracking scenario attribute maps as shown in~\Cref{fig:intro}. Note that frame rate takes the inverse in the maps. The detailed measurement metric is provided in the Supplementary Material.

Furthermore, we divide the attributes into two categories based on whether they damage motion or appearance, as shown by the background of the attribute map, \textit{i.e.}, motion and appearance-dominated methods may not perform well in the white and blue areas, respectively. 
In particular, pedestrian tracking scenarios with regular motion, high frame rate, and small movement amplitude, \emph{e.g.}, MOT17 and MOT20, are more motion-reliant; in tracking scenarios with highly complex motion patterns, \emph{e.g.} DanceTrack and SportsMOT, appearance is more effective than motion; BDD100K cannot provide reliable motion information due to the low frame rate and large movement amplitude. 
Our observations are consistent with how previous methods balance motion and appearance information. 
For exmaple, GHOST~\cite{seidenschwarz2023simple} sets the optimal weight between appearance and motion~(percentage of motion) as 0.6 for MOT17, 0.8 for MOT20, 0.4 for BDD100K, and 0.4 for DanceTrack. 
ByteTrack~\cite{bytetrack} utilizes motion to construct the affinity matrix in MOT17 and MOT20, and use appearance in BDD100K; it has very poor performance using motion on DanceTrack and SportsMOT in which the target motion is complex. For a tracker to have great generalizability, it is essential to take into account these attributes.

\begin{figure*}[t]
\centering
\includegraphics[width=0.95\linewidth]{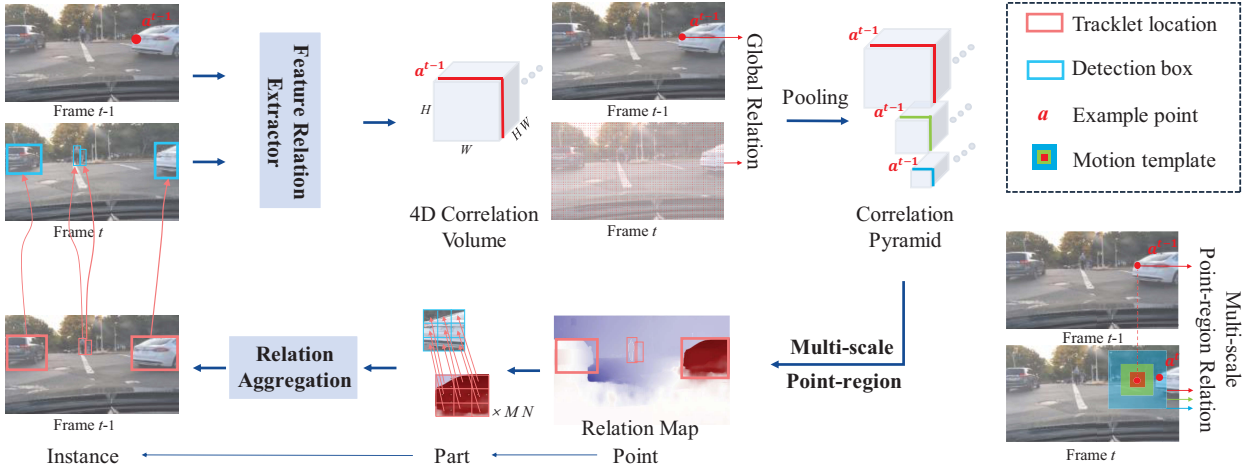}
\caption{{\bf Overview of our GeneralTrack.} The {\bf Feature Relation Extractor} obtains global dense relations with frame $t$ for each point in frame $t-1$ by a 4D correlation volume. Then by constructing a correlation pyramid, we transform the global relations into {\bf Multi-scale Point-region Relations}, and form a relation map for frame $t-1$. Finally, We progressively perform {\bf Relational Aggregation} to aggregate point-wise relation into instance-wise relation and achieve the association between tracklets and detections.}
\label{fig:Overview}
\end{figure*}

\subsection{Overview of GeneralTrack}
\noindent{\bf  Notation.}
For online video streaming, we first process the current frame~$t$ with YOLOX~\cite{ge2021yolox} to obtain the detection results. The detections are denoted as $\mathcal{D}^{t} = \{\mathbf{d}_{i}^{t}\}_{i=1}^N$ containing $N$ detections in frame~$t$, where $\mathbf{d}^{t}_{i}$ represents the position and size of a detection bounding box. We denote the set of $M$ tracklets by $\mathbb{T} = \{\mathcal{T}_{j}\}_{j=1}^M$. $\mathcal{T}_{j}$ is a tracklet with identity $j$ and is defined as $\mathcal{T}_{j}$ = \{$\mathbf{l}^{t_0}_{j}$, $\mathbf{l}^{t_0+1}_{j}$, ..., $\mathbf{l}^{t}_{j}$\}, where $\mathbf{l}^{t}_{j}$ is the location in frame~$t$, and $t_0$ is the initialized moment.

Our GeneralTrack follows the well-known tracking-by-detection paradigm~\cite{bytetrack}. Given the current frame~$t$, we obtain its detections $\mathcal{D}^{t}$ and the set of $M$ tracklets $\mathbb{T}$ up to frame~$t-1$. Then we associate existing tracklets $\mathbb{T}$ with current detections $\mathcal{D}^{t}$ by constructing the point-wise relations  between frame~$t-1$ and frame~$t$, and transforming them to the instance-wise associations. As shown in Figure~\ref{fig:Overview}, this process consists of three stages:
(i)  We use {\bf Feature Relation Extractor}~(\cref{Feature Relation Extractor}) to construct global dense relations with frame~$t$ for each point in frame~$t-1$ by a 4D correlation volume.
(ii)  Then we transform the global relations into {\bf Multi-scale Point-region Relations}~(\cref{ Multi-scale Point-region Relations}), and form a relation map for frame~$t-1$ in which each point represents its movement trends.
(iii)  Finally, we progressively perform {\bf Hierarchical Relational Aggregation}~(\cref{Relational Aggregation }) according to the \textit{point-part-instance} hierarchy to associate the tracklets and detections. 
All stages are differentiable and composed into an end-to-end trainable architecture.

\subsection{Feature Relation Extractor}
\label{Feature Relation Extractor}
Considering that the target could be very {\bf \textit{ small}} or {\bf \textit{ occluded}}, we exploit a extractor to capture the relationship at the point level.
Given a pair of consecutive RGB images, $\mathbf{I}^{t-1}$ and $\mathbf{I}^t$, a convolutional neural network  encodes them into two dense feature maps at a lower resolution, denoted as $\mathbf{F}^{t-1}$, $\mathbf{F}^{t} \in \mathbb{R}^{H \times W \times  D}$, where $H, W$ are respectively 1/8 of the image height and width, and $D$ is the feature dimension.

After obtaining the pair of consecutive feature maps, $\mathbf{F}^{t-1}$ and $\mathbf{F}^{t}$, we compute the global dense relations by constructing a full correlation volume between them. The correlation volume, $\mathbf{C}^\text{global}$, is formed by taking the dot product between all pairs of feature vectors as follows:
\begin{equation}
\begin{aligned}
	&\mathbf{C}^\text{global}\left(\mathbf{F}^{t-1},\mathbf{F}^{t}\right) \in \mathbb{R}^{H \times W \times H \times W}, \\
	&c_{i j k l}=\sum^\emph{D}_{d=1} f_{i j d}^{t-1} \cdot f_{k l d}^{t}, \\
\end{aligned}
\end{equation}
The element $c_{i j k l}$ in $\mathbf{C}^\text{global}$ represents the relation between the $(i, j)$-th feature point in frame~$t-1$ and the $(k, l)$-th feature point in frame~$t$. 

Instance-level features would be damaged when the target is too small or occluded, whereas point-wise relations between adjacent frames are robust in this scenario. 

\begin{figure*}[t]
\centering
\includegraphics[width=0.9\linewidth]{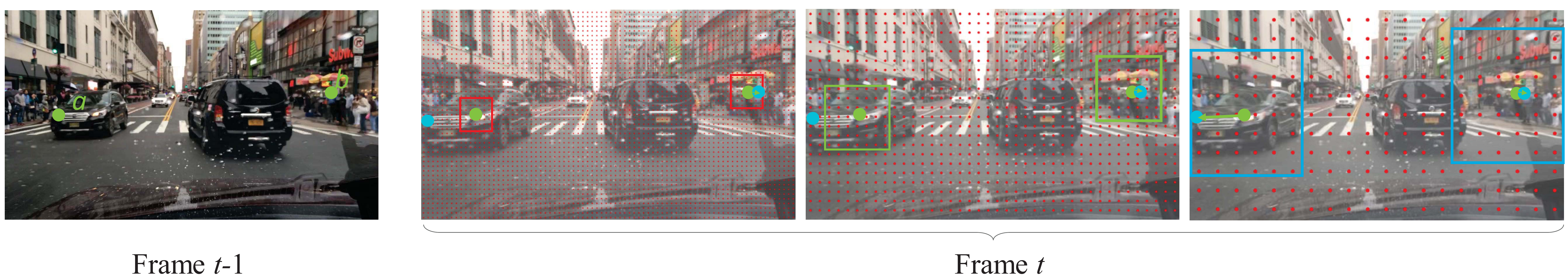}
\caption{{\bf Multi-scale point-region Relation} on correlation pyramid. With downsampling, the searching region becomes progressively larger~(red, green and blue box). Two examples points $a$ and $b$ are given in the figure, where the \textcolor{green}{green dot} and \textcolor{blue}{blue dot} represent the target point in frame $t-1$ and frame $t$ respectively. \textbf{(a)} Headlights of the car. The car moves very fast with large displacement, its relation point in frame $t$ is not captured until the layer with the largest scale \textbf{(b)} Head of the man. Due to the relatively small movement of people, its relation can be obtained at the first level of the pyramid~(the highest resolution). Such a relation searching paradigm can be flexibly adapted to both large and small displacements with low computational resources.
}
\label{fig:figure3}
\end{figure*}

\subsection{ Multi-scale Point-region Relation}
\label{ Multi-scale Point-region Relations}
This part is to adapt the tracker to {\bf \textit{ various frame rates}} as well as {\bf \textit{ the amplitude of position variation}}.
As demonstrated in Figure~\ref{fig:figure3}, there is a wide range of speeds among different classes, such as fast cars and slow people, and the same problem exists at different frame rates. 
Besides, $\mathbf{C}^\text{global}$ contains dense global relations, with significant invalid relations that would cause significant computational costs and slow convergence.
For tracker's flexibility and computational simplicity, we transform global relations into multi-scale point-region relations.

Inspired by the multi-scale 4D volume in~\cite{raft,dong2023rethinking}, we downsample the correlation volume and obtain the correlation pyramid $\{\mathbf{C}_s\}_{s=0}^S$ by pooling the last two dimensions as follows:
\begin{equation}
\begin{aligned}
	&\mathbf{C}_s = \mathtt{pooling}(\mathbf{C}_{\textit{s}-1}), \mathbf{C}_\textit{s} \in \mathbb{R}^{\textit{H} \times \textit{W} \times \textit{H}/2^\textit{s} \times \textit{W}/2^\textit{s}}, \\ \\
\end{aligned}
\end{equation}
where $s$ is the layer of pyramid and $\mathbf{C}_0$ is initialized as $\mathbf{C}^\text{global}$. The set of volumes provides relations from points in frame~$t-1$ to each area in frame~$t$ at different scales.

To reduce the invalid relations, we set a center-based relation searching region based on $\{\mathbf{C}_s\}_{s=0}^S$ as follows:
\begin{equation}
\begin{aligned}
	&\mathcal{P}\left(\mathbf{x}\right)_s=\left\{\mathbf{x}/2^s+\mathbf{r}  : \|\mathbf{r}\| \leq R\right\}, \\
\end{aligned}
\end{equation}
where $\mathbf{x}\in \mathbb{Z}^2$ is a coordinate in $\mathbf{F}^{t-1}$ and $\mathbf{r}\in \mathbb{Z}^2$ is the displacement from point $\mathbf{x}/2^s$ in the $s$-th layer in the correlation pyramid. The relation searching is restricted in the neighborhood of the coordinate $\mathbf{x}/2^s$, schematically as $\|\mathbf{r}\| \leq R$, \emph{i.e.} the maximum displacement in any direction is $R$. At each level of $\{\mathbf{C}_s\}_{s=0}^S$, the searching region in $\mathbf{F}^{t}$ gradually increases as the resolution decreases, as the red, green and blue boxes in Figure~\ref{fig:figure3}.

We employ $\mathcal{P}\left(\mathbf{x}\right)_s$ to search the relation on correlation pyramid $\{\mathbf{C}_s\}_{s=0}^S$ to form multi-scale point-region relation map $\mathbf{O} \in \mathbb{R}^{H \times W \times (S+1)(2\emph{R}+1)^2}$ as follows:
\begin{equation}
\begin{aligned}
	\mathbf{O}_s &= \mathtt{search}(\mathbf{C}_\textit{s}, \mathcal{P}\left(\mathbf{x}\right)_\textit{s}),  \\
	\mathbf{O}_{\ } &= \mathtt{concat}(\left\{\mathbf{O}_\textit{s}\right\}_{\textit{s}=0}^\textit{S}) ,
\end{aligned}
\end{equation}
where the function $\rm{search}$ means searching operation on each layer with $\mathcal{P}\left(\mathbf{x}\right)_s$. $\{\mathbf{O}_s \in \mathbb{R}^{H \times W \times (2\emph{R}+1)^2}\}_{s=0}^S$ is the point-region relations at each layer in $\{\mathbf{C}_s\}_{s=0}^S$, and they are concatenated at the last dimension. $\mathbf{O}$ contains the motion trends of the points in frame $t$-1 to frame t in all directions and at all ranges. 

In essence, the multi-scale point-region relation implicitly contains a vision and geometry based motion template, with different layers representing different displacement scales, and different points in the same layer representing different directions and amplitudes, which can be adapted to various frame rates as well as the amplitude of position variation flexibly.

\subsection{Hierarchical Relation Aggregation }
\label{Relational Aggregation }
In current frame $t$, the detection set $\mathcal{D}^{t}$ contains $N$ detections and the tracklets set $\mathbb{T}$ contains $M$ tracklets. The relation map $\mathbf{O}$ encodes frame~$t-1$'s point-wise relations to multi-scale regions in frame~$t$. Here we translate point-wise into instance-wise relations to construct tracklet-detection relation matrix $\mathbf{P}^{\rm{rela}}  \in \mathbb{R}^{N \times M}$ for data association.

To address the {\bf \textit{ shape variation of instances}}, we treat the instance as a flexible body.
Instead of treating the instance as a rigid body as traditional motion or appearance methods do, we progressively aggregate relations through a point-part-instance hierarchy.
For the $\emph{i}$-th detection $\mathbf{d}_{i}^{t}$ in $\mathcal{D}^{t}$ and the $\emph{j}$-th tracklet's location $\mathbf{l}_{j}^{t-1}$ in $\mathbb{T}$, we divide the instances $\mathbf{d}_{i}^{t}$ and $\mathbf{l}_{j}^{t-1}$ into $v \times v$ parts. We apply RoIAlign~\cite{he2017mask} to $\mathbf{l}_{j}^{t-1}$ on $\mathbf{O}$ and obtain the  $\emph{j}$-th tracklet's part-wise relation $\mathbf{O}_{j}  \in \mathbb{R}^{v \times v \times D}$, where $v \times v$ is the RoIAlign size and $D$ is the feature dimension. Then we encode the relative position of each part between $\mathbf{d}_{i}^{t}$ and $\mathbf{l}_{j}^{t-1}$, denotes as $\mathbf{E}_{ij}  \in \mathbb{R}^{v \times v \times 2}$, where its element represents the displacement of the centroids of corresponding parts. Then we concatenate $\mathbf{O}_{j}$ and $\mathbf{E}_{ij}$ and apply the convolution operation to it.  Finally, we obtain the prediction score with a multi-layer perception~(MLP) as follows:
\begin{equation}
\begin{aligned}
 	&p_{ij}^{\rm{rela}} = \mathtt{MLP}\left(\mathtt{conv}(\mathtt{concat}\left(\mathbf{O}_{\emph{j}},\mathbf{E}_{\emph{ij}} \right))\right).
\end{aligned}
\end{equation}

Eventually, we use the Hungarian Algorithm~\cite{hungarian} for $\mathbf{P}^{\rm{rela}}$ and complete the association. 
For the lost tracklets, we use Kalman filter to retrieve them.

\begin{table*}[t]
\centering
\renewcommand{\arraystretch}{1}
\resizebox{0.95\linewidth}{!}{
\setlength{\tabcolsep}{0.9em}%
	\begin{tabular}{lcccccccccc}
\hline
                                      & \textbf{Venue} & \textbf{mHOTA$\uparrow$} & \textbf{mIDF1$\uparrow$} & \textbf{mMOTA$\uparrow$} & \textbf{HOTA$\uparrow$} & \textbf{IDF1$\uparrow$} & \textbf{MOTA$\uparrow$} & \textbf{IDs$\downarrow$} & \textbf{MT$\uparrow$}   & \textbf{ML$\downarrow$} \\ \hline
\textbf{\textit{validation}}          &                &                          &                          &                          &                         &                         &                         &                          &                         &                         \\
QDTrack~\cite{quasi}                  & CVPR'21        & -                        & 50.8                     & 36.6                     & -                       & 71.5                    & 63.5                    & \bf{6262}                & 9481                    & 3034                    \\
Unicorn~\cite{yan2022towards}         & ECCV'22        & -                        & 54.0                     & 41.2                     & -                       & 71.3                    & 66.6                    & 10876                    & \textcolor{blue}{10296} & \textcolor{blue}{2505}  \\
MOTR~\cite{zeng2022motr}              & ECCV'22        & -                        & 44.8                     & 32.3                     & -                       & 65.8                    & 56.2                    & -                        & -                       & -                       \\
TETer~\cite{li2022tracking}           & ECCV'22        & -                        & 53.3                     & 39.1                     & -                       & -                       & -                       & -                        & -                       & -                       \\
ByteTrack~\cite{bytetrack}            & ECCV'22        & 45.3                     & 54.8                     & 45.2                     & 61.3                    & 70.4                    & \bf{69.1}               & 9140                     & 9626                    & 3005                    \\
MOTRv2~\cite{motrv2}                  & CVPR'23        & -                        & \bf{56.5}                & 43.6                     & -                       & \bf{72.7}               & 65.6                    & -                        & -                       & -                       \\
GHOST~\cite{seidenschwarz2023simple}  & CVPR'23        & \textcolor{blue}{45.7}   & 55.6                     & \textcolor{blue}{44.9}   & \textcolor{blue}{61.7}  & 70.9                    & 68.1                    & -                        & -                       & -                       \\
\rowcolor[HTML]{EFEFEF} GeneralTrack(Ours) &                & \bf{46.9}                & \textcolor{blue}{56.2}   & \bf{46.4}                & \bf{63.1}               & \bf{72.7}               & \textcolor{blue}{68.8}  & 8496                     & \bf{11830}              & \bf{2035}               \\ \hline
\textbf{\textit{test}}                &                &                          &                          &                          &                         &                         &                         &                          &                         &                         \\
DeepBlueAI                            & -              & -                        & 38.7                     & 31.6                     & -                       & 56.0                    & 56.9                    & 25186                    & 10296                   & 12266                   \\
madamada                              & -              & -                        & 43.0                     & 33.6                     & -                       & 55.7                    & 59.8                    & 42901                    & 16774                   & \textcolor{blue}{5004}  \\
QDTrack~\cite{bdd100k}                & CVPR'21        & 41.9                     & 52.4                     & 35.7                     & 60.5                    & \textcolor{blue}{72.5}  & 64.6                    & \bf{10790}               & 17353                   & 5167                    \\
ByteTrack~\cite{bytetrack}            & ECCV'22        & -                        & 55.8                     & \bf{40.1}                & -                       & 71.3                    & \bf{69.6}               & 15466                    & \textcolor{blue}{18057} & 5107                    \\
GHOST~\cite{seidenschwarz2023simple}  & CVPR'23        & \textcolor{blue}{46.8}   & \bf{57.0}                & 39.5                     & \textcolor{blue}{62.2}  & 72.0                    & 68.9                    & -                        & -                       & -                       \\
\rowcolor[HTML]{EFEFEF} GeneralTrack(Ours) &                & \bf{47.9}                & \textcolor{blue}{56.9}   & \textcolor{blue}{39.9}   & \bf{63.7}               & \bf{73.6}               & \textcolor{blue}{69.1}  & \textcolor{blue}{14489}  & \bf{21281}              & \bf{3715}               \\ \hline
\end{tabular}
	
}
\caption{Comparison with the state-of-the-art methods on BDD100K. The two best results for each metric are highlighted in bolded and blue. Note that some methods only provide the \textit{validation set} result without \textit{test set}.}  
\label{table1}  
\end{table*}

\begin{table}[t]
\centering
\renewcommand{\arraystretch}{1.0}
\setlength{\tabcolsep}{4.5pt}
\Huge
\resizebox{0.95\linewidth}{!}{
\begin{tabular}{lcccccc}
\hline
                                        & \textbf{Venue} & \textbf{HOTA$\uparrow$} & \textbf{MOTA$\uparrow$} & \textbf{IDF1$\uparrow$} & \textbf{AssA$\uparrow$} & \textbf{DetA$\uparrow$} \\ \hline
GTR~\cite{GTR}                          & CVPR'22        & 54.5                    & 67.9                    & 55.8                    & 45.9                    & 64.8                    \\
ByteTrack~\cite{bytetrack}              & ECCV'22        & 64.1                    & 95.9                    & 71.4                    & 52.3                    & 78.5                    \\
OC-SORT~\cite{ocsort}                   & CVPR'23        & 73.7                    & \textcolor{blue}{96.5}  & 74.0                    & 61.5                    & \textcolor{blue}{88.5}  \\
MixSort-Byte*~\cite{sportsmot}          & ICCV'23        & \textcolor{blue}{65.7}  & 96.2                    & 74.1                    & 54.8                    & 78.8                    \\
MixSort-OC*~\cite{sportsmot}            & ICCV'23        & \bf{74.1}               & \textcolor{blue}{96.5}  & \textcolor{blue}{74.4}  & \bf{62.0}               & \textcolor{blue}{88.5}  \\
\rowcolor[HTML]{EFEFEF} GeneralTrack(Ours) & -              & \bf{74.1}               & \bf{96.8}               & \bf{76.4}               & \textcolor{blue}{61.7}  & \bf{89.0}               \\ \hline
\end{tabular}
}
\caption{Comparison with the state-of-the-art methods on SportsMOT. The two best results for each metric are highlighted in bolded and blue. Note that $*$ denotes the methods use the validation set for training the association model.}  
\label{table2}  
\end{table}

\begin{table}[t]
\centering
\renewcommand{\arraystretch}{1.0}
\setlength{\tabcolsep}{4.5pt}
\Huge
\resizebox{0.95\linewidth}{!}{
\begin{tabular}{lcccccc}
\hline
                                                       & \textbf{Venue}                              & \textbf{HOTA$\uparrow$}                     & \textbf{MOTA$\uparrow$}                     & \textbf{IDF1$\uparrow$}                     & \textbf{AssA$\uparrow$}                     & \textbf{DetA$\uparrow$}                     \\ \hline
\textbf{\textit{Transformer based:}}                   &                                             &                                             &                                             &                                             &                                             &                                             \\
MOTR~\cite{zeng2022motr}                               & ECCV'22                                     & 54.2                                        & 79.7                                        & 51.5                                        & 40.2                                        & 73.5                                        \\ \hline
\textbf{\textit{Hybird based:}} & \multicolumn{1}{l}{{\color[HTML]{9B9B9B} }} & \multicolumn{1}{l}{{\color[HTML]{9B9B9B} }} & \multicolumn{1}{l}{{\color[HTML]{9B9B9B} }} & \multicolumn{1}{l}{{\color[HTML]{9B9B9B} }} & \multicolumn{1}{l}{{\color[HTML]{9B9B9B} }} & \multicolumn{1}{l}{{\color[HTML]{9B9B9B} }} \\
MOTRv2~\cite{motrv2}          & CVPR'23              & 69.9                 &  91.9                 &  71.7                 &  59.0                 &  83.0                 \\ \hline
\textbf{\textit{CNN based:}}                           & \multicolumn{1}{l}{}                        & \multicolumn{1}{l}{}                        & \multicolumn{1}{l}{}                        & \multicolumn{1}{l}{}                        & \multicolumn{1}{l}{}                        & \multicolumn{1}{l}{}                        \\
ByteTrack~\cite{bytetrack}                             & ECCV'22                                     & 47.7                                        & 89.6                                        & 53.9                                        & 32.1                                        & 71.0                                        \\
FineTrack~\cite{focus}                                 & CVPR'23                                     & 52.7                                        & 89.9                                        & \bf{59.8}                                   & 38.5                                        & 72.4                                        \\
OC-SORT~\cite{ocsort}                                  & CVPR'23                                     & 55.1                                        & \bf{92.2}                                   & 54.9                                        & \textcolor{blue}{40.4}                      & 80.4                                        \\
GHOST~\cite{seidenschwarz2023simple}                   & CVPR'23                                     & \textcolor{blue}{56.7}                      & 91.3                                        & 57.7                                        & 39.8                                        & \textcolor{blue}{81.1}                      \\
\rowcolor[HTML]{EFEFEF} GeneralTrack~(Ours)                        & -                                           & \bf{59.2}                                   & \textcolor{blue}{91.8}                      & \textcolor{blue}{59.7}                      & \bf{42.8}                                   & \bf{82.0}                                   \\ \hline
\end{tabular}
}
\caption{Comparison with the state-of-the-art methods on DanceTrack. The two best results are highlighted in bolded and blue. Note that MOTRv2 uses both YOLOX and MOTR with more than two hundred times training resource usage than ours.}  
\label{table3}  
\end{table}

\subsection{Training}
\label{Training}
We pick two consecutive frames in the video as a training sample. Two raw images are used as inputs $\mathbf{I}^{t-1}$ and $\mathbf{I}^t$ in the forward propagation process, and the grounding truth boxes are used as $\mathbb{T}$ and $\mathcal{D}^{t}$ respectively. The targets in two frames are combined in pairs to predict tracklet-detection relation $\mathbf{P}^{\rm{rela}}$ and we label positive or negative by whether they have the same identity.  Then, we supervise it with a weighted binary cross-entropy loss function~(weighted BCE Loss):
\begin{equation}
\begin{aligned}
	\mathcal{L}^{\text{wBCE}}&=\frac{1}{NM} \sum_{i=1}^{N}\sum_{j=1}^{M}-[w \cdot y_{ij} \log (p^{\rm{rela}}_{ij})  \\
	&+(1-y_{ij})\log(1-p^{\rm{rela}}_{ij})],
\end{aligned}
\end{equation}
where $p^{\rm{rela}}_{ij}$ denotes the predicted correlation score, and $y_{ij}$ indicates the ground truth correlation label, in which 1 and 0 represent the positive and negative correlation, respectively. There is a great difference between the amount of positive and negative samples, inspired by F³Net~\cite{f3net}, we add weights to the positive samples, \emph{i.e.}, the weighting factor $w$. To enhance the capability of Feature Relation Extractor, we use~\cite{kitti} for pre-training in the settings of dense flow and correspondence tasks.

\subsection{Multi-class Object Tracking}
\label{Multi-catagory Object Tracking}
FairMOT~\cite{fairmot} regards each tracked object as a class and associates the detection results by feature similarity. 
Some TbD methods, such as GHOST~\cite{seidenschwarz2023simple} and ByteTrack~\cite{bytetrack}, 
restrict the tracking process in each class. 
Target tracking can easily be interrupted due to misjudgments of the class by object detection. 
In contrast, we first conduct class-agnostic association, allowing objects detected as different classes across frames to be associated. Second, we determine the ``true class" as the most frequent class in the historical trajectory. Finally, we treat detections in the trajectory that differ from the ``true class" as false positives and categorize them into the ``true class".

\begin{table}[t]
\centering
\renewcommand{\arraystretch}{1.0}
\setlength{\tabcolsep}{2.5pt}
\Huge
\resizebox{0.95\linewidth}{!}{	
\begin{tabular}{lccccccc}
\hline
                                                                & \textbf{Venue} & \textbf{HOTA$\uparrow$} & \textbf{MOTA$\uparrow$} & \textbf{IDF1$\uparrow$} & \textbf{AssA$\uparrow$} & \textbf{DetA$\uparrow$} & \textbf{IDs$\downarrow$} \\ \hline
\textbf{\textit{MOT17}}                                         &                &                         &                         &                         &                         &                         &                          \\

MOTR~\cite{zeng2022motr}                                        & ECCV'22        & 57.8                    & 73.4                    & 68.6                    & 55.7                    & 60.3                    & 2439                     \\
ByteTrack~\cite{bytetrack}                                      & ECCV'22        & 63.1                    & \textcolor{blue}{80.3}& 77.3                    & 62.0                    & \textcolor{blue}{64.5}& 2196                     \\
OC-SORT~\cite{ocsort}                                           & CVPR'23        & \textcolor{blue}{63.2}& 78.0                    & \textcolor{blue}{77.5}& \bf{63.2}               & 63.2                    & \textcolor{blue}{1950}\\
MOTRv2~\cite{motrv2}                                            & CVPR'23        & 62.0                    & 78.6                    & 75.0                    & 60.6                    & 63.8                    & -                        \\
GHOST~\cite{seidenschwarz2023simple}                            & CVPR'23        & 62.8                    & 78.7                    & 77.1                    & -                       & -                       & 2325                     \\
\rowcolor[HTML]{EFEFEF}   \rowcolor[HTML]{EFEFEF} GeneralTrack(Ours) & -              & \bf{64.0}               & \bf{80.6}               & \bf{78.3}               & \textcolor{blue}{63.1}& \bf{65.1}               & \bf{1563}                \\

\hline
\textbf{\textit{MOT20}}                                         &                &                         &                         &                         &                         &                         &                          \\

ByteTrack~\cite{bytetrack}                                      & ECCV'22        & 61.3                    & \bf{77.8}               & \textcolor{blue}{75.2}& \textcolor{blue}{59.6}& \textcolor{blue}{63.4}& \textcolor{blue}{1223}\\
OC-SORT~\cite{ocsort}                                           & CVPR'23        & \bf{62.1}               & 75.5                    & \bf{75.9}               & \bf{62.0}               & -                       & \bf{913}                 \\
MOTRv2~\cite{motrv2}                                            & CVPR'23        & 60.3                    & 76.2                    & 72.2                    & 58.1                    & 62.9                    & -                        \\
GHOST~\cite{seidenschwarz2023simple}                            & CVPR'23        & 61.2                    & 73.7                    & \textcolor{blue}{75.2}& -                       & -                       & 1264                     \\
\rowcolor[HTML]{EFEFEF}   \rowcolor[HTML]{EFEFEF} GeneralTrack(Ours) & -              & \textcolor{blue}{61.4}& \textcolor{blue}{77.2}& 74.0                    & 59.5                    & \bf{63.7}               & 1627                     \\ 
\hline
\end{tabular}	
}
\caption{Comparison with the state-of-the-art methods on MOT17, MOT20. The two best results are highlighted in bolded and blue.}  
\label{table4}  
\end{table}

\section{Experiments}
\label{sec:Experiments}
\subsection{Setting}
\noindent{\bf Datasets and Metrics.} 
We evaluate GeneralTrack on BDD100K~\cite{bdd100k}, SportsMOT~\cite{sportsmot}, DanceTrack~\cite{dancetrack}, MOT17~\cite{mot16} and MOT20~\cite{mot20} datasets. As standard protocols, CLEAR MOT Metrics~\cite{clear-metric} and HOTA~\cite{luiten2021hota} are used for evaluation. For multi-class tracking, average metrics across all classes are added as well as Tracking-Every-Thing Accuracy~(TETA)~\cite{li2022tracking} metric for ranking as they did in BDD100K MOT Challenge 2023.

\noindent{\bf Implementation Details.}
For fair comparisons, we directly apply the publicly available detector of YOLOX~\cite{ge2021yolox}, trained by~\cite{seidenschwarz2023simple} for BDD100K, \cite{sportsmot} for SportsMOT, \cite{dancetrack} for DanceTrack, \cite{bytetrack} for MOT17, MOT20. Note that all the detection results we used are the same as most of the SOTAs compared in the tables. The number of downsampling in correlation pyramid $S$ is 3, the relation searching radius $R$ is 4. The RoIAlign size $v*v$ is set to 3 * 3 for DanceTrack, SportsMOT and 2 * 2 for BDD100K, MOT17, MOT20.

\begin{table}[t]
\centering
\renewcommand{\arraystretch}{1.0}
\setlength{\tabcolsep}{0.1pt}
\Huge
\resizebox{0.95\linewidth}{!}{
\begin{tabular}{c|cc|ccccccc}
\toprule
Setting & \textbf{MR}                               & \textbf{HRA}                              & mHOTA$\uparrow$ & mIDF1$\uparrow$ & mMOTA$\uparrow$ & HOTA$\uparrow$ & IDF1$\uparrow$ & MOTA$\uparrow$ & IDs$\downarrow$ \\  \midrule
\textit{\#1}& \Checkmark & \Checkmark & 47.1                     & 56.1                     & 46.1                     & 63.4                    & 72.5                    & 68.3                    & 8503                     \\
\textit{\#2}& \Checkmark & \Checkmark & 46.2                     & 54.5                     & 43.6                     & 62.4                    & 71.3                    & 66.0                    & 9070                     \\
\textit{\#3}& \Checkmark & \Checkmark & 42.9                     & 49.2                     & 37.5                     & 57.8                    & 63.9                    & 59.2                    & 11584                    \\ \hline
\textit{\#1}&            & \Checkmark & 46.9                     & 55.7                     & 45.6                     & 63.1                    & 72.2                    & 67.9                    & 9447                     \\
\textit{\#2}&            & \Checkmark & 45.3                     & 53.3                     & 42.2                     & 61.7                    & 70.1                    & 65.0                    & 10673                    \\
\textit{\#3}&            & \Checkmark & 41.7                     & 47.8                     & 36.3                     & 55.8                    & 61.2                    & 56.7                    & 14015                    \\  \midrule
\textit{\#1}& \Checkmark                                &                                           & 46.7                     & 55.5                     & 45.6                     & 62.8                    & 71.8                    & 67.9                    & 9070                     \\  \midrule
\multirow{4}{*}{\textit{\#1}}& \Checkmark                                & \Checkmark                                &                          &                          &                          &                         &                         &                         &                          \\
                                                                        & \multicolumn{2}{c|}{\textbf{SR}=1}                                                             & 46.9                     & 55.7                     & 46.0                     & 63.1                    & 72.1                    & 68.2                    & 9173                     \\
                                                                        & \multicolumn{2}{c|}{\textbf{SR}=4}                                                             & 47.1                     & 56.1                     & 46.1                     & 63.4                    & 72.5                    & 68.3                    & 8503                     \\
                                                                        & \multicolumn{2}{c|}{\textbf{SR}=7}                                                             & 46.9                     & 55.7                     & 46.0                     & 63.5                    & 72.7                    & 68.3                    & 8454                     \\ \bottomrule
\end{tabular}
}
\caption{Ablation study of Multi-scale Relation~(MR), Searching Radius~(SR), Hierarchical Relation Agregation~(HRA). MR is assessed under three frame rate settings, where $\textit{\#1}$, $\textit{\#2}$ and $\textit{\#3}$ represent the original as well as double and quadruple downsampling.}  
\label{table5}  
\end{table}

\subsection{Benchmark Evaluation}
To validate generalizability, we conduct experiments on five public benchmarks with significantly different characteristics: 
multi-class tracking in autonomous driving scenario BDD100K, accompanied by low frame rates, large variations and the presence of small targets~(in~\Cref{table1}); ball game scenario SportsMOT with complex motion patterns~(in~\Cref{table2}); dancing scenario DanceTrack, comes with serious motion complexity and large variation amplitude~(in~\Cref{table3}); pedestrian tracking scenarios MOT17 and MOT20 with regular motion, high frame rate, along with higher target densities and the presence of a large number of small targets~(in~\Cref{table4}).

\noindent{\bf Generalizability Analysis.}
Rather than requiring strong hypotheses and huge training resources, our approach can be generalized to diverse scenarios with great performance.
Even though ByteTrack selects different association information in different datasets and GHOST adjusts the weighting weights for association information in different datasets, we still outperform them on all datasets. 
For MOTRv2, which requires more than 200 times training resource usage than us, we perform better on all datasets except DanceTrack with very little training resources and time.  
More detailed analyses on benchmarks are provided in the supplementary material.

\begin{table}[t]
\centering
\renewcommand{\arraystretch}{1.21}
\Huge
\resizebox{0.95\linewidth}{!}{
\setlength{\tabcolsep}{0.5em}%
	
	\begin{tabular}{l|cccr}
	\toprule
		Class  & HOTA$\uparrow$& IDF1$\uparrow$& MOTA$\uparrow$& \multicolumn{1}{c}{IDs$\downarrow$} \\ \midrule 
		Pedestrian      & 50.3(+0.1)& 60.7(+0.1)& 55.6(+0.2)& 2236($\downarrow~~$1.2\%)\\
		Rider           & 43.7(\textbf{+3.6})& 57.9(\textbf{+2.8})& 46.3(\textbf{+2.6})& 52($\downarrow$\textbf{44.2\%})\\
		Car             & 66.2(+0.0)& 75.4(+0.1)& 73.1(-0.1)& 6018($\downarrow$~~~1.6\%)\\
		Bus             & 60.0(\textbf{+2.1})& 69.1(\textbf{+1.8})& 56.5(\textbf{+1.9})& 70($\downarrow$\textbf{35.7\%})\\
		Truck           & 54.2(+0.8)& 61.7(\textbf{-1.1})& 48.7(\textbf{-1.5})& 219($\downarrow$\textbf{12.3\%})\\
		Train           & 0.0~~(+0.0)& 0.0~~(+0.0)& -0.6~(+0.0)& 0.0($\downarrow$~~~0.0\%)\\
		Motorcycle      & 46.6(\textbf{+1.0})& 58.6(+0.7)& 39.2(\textbf{+3.2})& 11($\downarrow$\textbf{27.3\%})\\
		Bycicle         & 47.8(+0.2)& 60.1(+0.3)& 43.1(+0.5)& 144($\downarrow$~~~0.7\%)\\ \midrule 
		Detect\_average & 63.3(+0.1)& 72.5(+0.0)& 68.4(-0.1)& 8750($\downarrow$~~~2.8\%)\\
		Class\_average  & 46.1(\textbf{+1.0})& 55.5(+0.6)& 45.2(+0.9)& 8750($\downarrow$~~~2.8\%)\\ \bottomrule
	\end{tabular}

}
\caption{Ablation study of Class Relaxation and Correction in multi-calss tracking, the performance changes they bring are given in parentheses, with the larger changes highlighted in bold.}  
\label{table6}  
\end{table}

\begin{figure*}[t]
\centering
\includegraphics[width=0.95\linewidth]{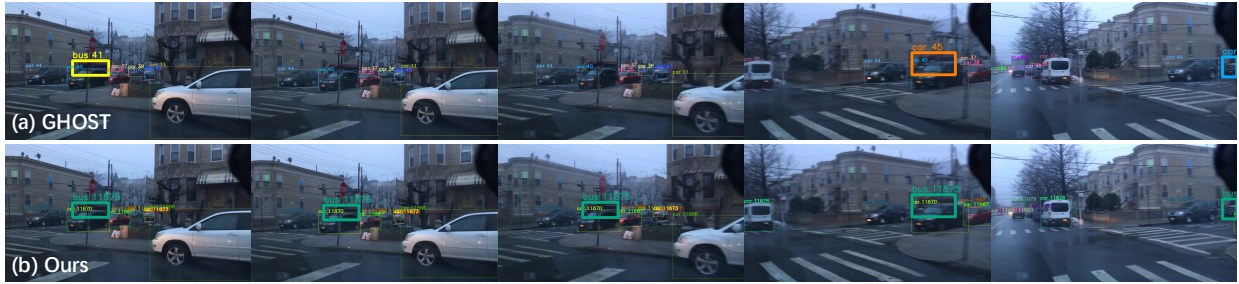}
\caption{{\bf Visualization of tracking results comparison.} Note that we use exactly the same detection results as for GHOST. The boxes of different colors represent the bounding boxes with different identities. The red bus shown in bold is the target of our comparison. It experienced tracklet interruptions, id switch, and misclassification in GHOST, in the meantime these were resolved in our approach.
}
\label{fig:figure4}
\end{figure*}

\subsection{Ablation Studies}
We conduct ablation experiments on the BDD100K validation set to evaluate our design of each component. Note that the ablation results are evaluated using the toolkit and the benchmark results are obtained by submitting to the evaluation server.

\noindent{\bf Multi-scale Relation.}
Multi-scale relation~(MR) is designed to enable the tracker to cope with different frame rates and variation amplitudes. We conducted experiments at three frame rate settings~(original frame rate as well as double and quadruple downsampling). 
As shown in the first six rows of~\Cref{table5}, all metrics got worse with ablation under all three settings, indicating the key role MR plays in tracking. As the frame rate drops, the performance variations intensify, and under setting3, MR brings a boost~(+2.0 HOTA, +2.7 IDF1, +2.5 MOTA, +1.2 mHOTA, +1.4 mIDF1, +1.2 mMOTA). This suggests that the lower the frame rate, the more powerful MR is in the scene.

\noindent{\bf Searching Radius.}
As illustrated in~\Cref{table5}, changing the searching radius has little effect on tracker's performance. Because the correlation pyramid is robust to this parameter, the correlation pyramid has the flexibility to cover multi-scale regions at any radius.

\noindent{\bf Hierarchical Relation Agregation.}
We ablate this part by setting the RoiAlign parameter $v*v$ to $1*1$ and treating it as a rigid body as in traditional methods. As~\Cref{table5} shows, all metrics worsened with ablation, suggesting that this component affects the tracking of all targets. This highlights the importance of transforming a rigid body into a flexible body during the matching procedure.

\noindent{\bf Class Relaxation and Correction.}
In order to observe the performance changes in each class, we list them separately in~\Cref{table6}. The performance of rider, bus, and motorcycle improved substantially, reflecting that many targets are mistaken for other classes during the detection process, resulting in unsuccessful matches. Although two metrics of truck decreases, the average metrics still improve, suggesting that our method reduces a large number of false negatives while having a low false positive rate. This compensates for the weakness of the detector's classification ability.

\subsection{Visualization}
We focus on the red bus in Figure~\ref{fig:figure4}, where GHOST~\cite{seidenschwarz2023simple} experienced tracklet interruptions, id switch, and misclassification. We investigate why these phenomena occur. It is because low score detection cannot depend on a valid Intersection over Union~(IoU) or a poor appearance to be tracked when there is a large movement amplitude; the misclassification and id switch are due to the detector's class misjudgment. In contrast, we performed a complete track of it and successfully corrected the classification errors.

\begin{table}[t]
\centering
\renewcommand{\arraystretch}{1.1}
\setlength{\tabcolsep}{4pt}
\Huge
\resizebox{0.9\linewidth}{!}{
	\setlength{\tabcolsep}{0.3em}%
	
\begin{tabular}{ccccccccc}
\toprule
Class                                & Car                                  & Peds & Rider & Bus  & Truck           & Train & Motocy & Bycicle \\ \midrule 
\multicolumn{1}{c|}{Setting}         & \multicolumn{8}{c}{\textbf{Source} \& \textbf{Target}}                                                            \\ \midrule
\multicolumn{1}{c|}{HOTA$\uparrow$}  & 66.2                                 & 50.4       & 47.3  & 62.1 & 55.0            & 0     & 47.6       & 48.0    \\
\multicolumn{1}{c|}{IDF1$\uparrow$}  & 75.7                                 & 60.8       & 60.7  & 70.9 & 60.6            & 0     & 59.3       & 60.4    \\
\multicolumn{1}{c|}{MOTA$\uparrow$}  & 73.0                                 & 55.8       & 48.9  & 58.4 & 47.2            & -0.6  & 42.4       & 43.6    \\
\multicolumn{1}{c|}{IDs$\downarrow$} & 5917                                 & 2209       & 29    & 45   & 192             & 0     & 8          & 143     \\ \midrule 
\multicolumn{1}{c|}{Setting}         & \multicolumn{1}{c|}{\textbf{Source}} &            &       &      & \textbf{Target} &       &            &         \\ \midrule 
\multicolumn{1}{c|}{HOTA$\uparrow$}  & \multicolumn{1}{c|}{65.8}            & 48.9       & 45.6  & 61.8 & 54.6            & 0     & 47.7       & 47.7    \\
\multicolumn{1}{c|}{IDF1$\uparrow$}  & \multicolumn{1}{c|}{74.9}            & 58.6       & 57.7  & 70.4 & 60.4            & 0     & 60         & 59.8    \\
\multicolumn{1}{c|}{MOTA$\uparrow$}  & \multicolumn{1}{c|}{72.8}            & 54.3       & 44.1  & 58.9 & 46.9            & -0.6  & 41.7       & 43.3    \\
\multicolumn{1}{c|}{IDs$\downarrow$} & \multicolumn{1}{c|}{6186}            & 2790       & 23    & 44   & 140             & 0     & 8          & 152     \\ \bottomrule
\end{tabular}

}
\caption{Domain generalization for data with different classes. }  
\label{table7}  
\end{table}

\begin{table}[t]
\centering
\renewcommand{\arraystretch}{1.12}
\Huge
\resizebox{0.9\linewidth}{!}{
	\setlength{\tabcolsep}{0.3em}%
	
	\begin{tabular}{cc|ccccc}
		\toprule
		\begin{tabular}[c]{@{}c@{}}Training\\ \textbf{(Source)}\end{tabular} & \begin{tabular}[c]{@{}c@{}}Inference\\ \textbf{(Target)}\end{tabular} & HOTA$\uparrow$ & MOTA$\uparrow$ & IDF1$\uparrow$  & AssA$\uparrow$ &DetA$\uparrow$ \\ \midrule 
		SportsMOT                                                           & SportsMOT                                                            & 75.0 & 95.6 & 77.9 & 63.6 & 88.4 \\
		BDD100K                                                             & SportsMOT                                                            & 73.8 & 95.7 & 76.7  & 61.6 & 88.4 \\ \midrule
		DanceTrack                                                          & DanceTrack                                                           & 56.9 & 90.1 & 57.5  & 41.1 & 79.1 \\
		BDD100K                                                             & DanceTrack                                                           & 54.9 & 89.2 & 55.3  & 38.4 & 78.7 \\ \bottomrule
	\end{tabular}
	
}
\caption{Domain generalization for data in different datasets. }  
\label{table8}  
\end{table}

\subsection{Domain Generalization}
In this part, we take a further step to experiment and analyze whether our GeneralTrack can also perform well in domain generalization setting, which is a new and critical challenge in MOT field~\cite{segu2023darth}. Domain generalization in MOT is organized into two phases, detection and association. Here we conduct domain generalization experiments for the association part, \emph{i.e.}, training only on the source domain and inference on the unseen target domain without fine-tuning.

\noindent{\bf Cross-class and Cross-dataset Experiments.}
In~\Cref{table7}, we set up domain generalization between different classes, \emph{i.e.}, we train only with targets in car class and inference over all classes. It can be noted that there is still excellent tracking performance on the seven unseen classes. 
As shown in~\Cref{table8}, we train on BDD100K and then generalize to SportsMOT and DanceTrack without fine-tuning. Comparison with results trained and inferred in the same domain demonstrates that our GeneralTrack has strong domain generalization capabilities.

\noindent{\bf Analysis.}
Focusing on the local key texture of the target is more generalizable than the global structural information~\cite{ng2003sift, hou2021towards}. We accomplish tracking by constructing a point-wise relations between frames, which is based on low-level visual information such as textures, shapes, and corners points, \textit{etc.}. These low-level visual information is shared by all targets and has greater flexibility and generalizability. So these factors enable our GeneralTrack the ability to domain generalization.

\section{Conclusion}
\label{sec:Conclusion}
In this paper, we explore the difficulties in trackers' generalizability to diverse scenarios, and concretize them into a set of tracking scenario attributes that can guide the design of future trackers. 
Furthermore, guided by these attributes, we propose a “point-wise to instance-wise relation” framework for MOT, \textit{i.e.}, GeneralTrack. We achieve excellent performance on multiple datasets while avoiding the need to balance motion and appearance and experimentally demonstrated great potential for domain generalization with unseen data distributions~(cross-dataset, cross-class). 

\noindent \textbf{Limitation and Future Work.} We focus more on modeling inter-frame relations and do not extend it to cross-frame relations. Inspired by~\cite{wang2023omnimotion,zheng2023pointodyssey}, in our next version, we will construct the relations between multiple frames based on video clips to achieve better tracking performance.

\section*{Acknowledgement}
This work was supported in part by NSFC under Grants 62088102 and 62106192, Natural Science Foundation of Shaanxi Province under Grant 2022JC-41, and Fundamental Research Funds for the Central Universities under Grant XTR042021005.

{
    \small
    \bibliographystyle{ieeenat_fullname}
    \bibliography{main}

\begin{thebibliography}{66}
\providecommand{\natexlab}[1]{#1}
\providecommand{\url}[1]{\texttt{#1}}
\expandafter\ifx\csname urlstyle\endcsname\relax
  \providecommand{\doi}[1]{doi: #1}\else
  \providecommand{\doi}{doi: \begingroup \urlstyle{rm}\Url}\fi

\bibitem[Aharon et~al.(2022)Aharon, Orfaig, and Bobrovsky]{botsort}
Nir Aharon, Roy Orfaig, and Ben-Zion Bobrovsky.
\newblock Bot-sort: Robust associations multi-pedestrian tracking.
\newblock \emph{arXiv preprinbytetrack arXiv:2206.14651}, 2022.

\bibitem[Andriluka et~al.(2008)Andriluka, Roth, and Schiele]{tracking-by-detection}
Mykhaylo Andriluka, Stefan Roth, and Bernt Schiele.
\newblock People-tracking-by-detection and people-detection-by-tracking.
\newblock In \emph{CVPR}, pages 1--8, 2008.

\bibitem[Bergmann et~al.(2019)Bergmann, Meinhardt, and Leal-Taixe]{tracktor}
Philipp Bergmann, Tim Meinhardt, and Laura Leal-Taixe.
\newblock Tracking without bells and whistles.
\newblock In \emph{ICCV}, pages 941--951, 2019.

\bibitem[Bernardin and Stiefelhagen(2008)]{clear-metric}
Keni Bernardin and Rainer Stiefelhagen.
\newblock Evaluating multiple object tracking performance: the clear mot metrics.
\newblock \emph{JIVP}, 2008:\penalty0 1--10, 2008.

\bibitem[Bewley et~al.(2016)Bewley, Ge, Ott, Ramos, and Upcroft]{sort}
Alex Bewley, Zongyuan Ge, Lionel Ott, Fabio Ramos, and Ben Upcroft.
\newblock Simple online and realtime tracking.
\newblock In \emph{ICIP}, pages 3464--3468, 2016.

\bibitem[Cao et~al.(2023)Cao, Pang, Weng, Khirodkar, and Kitani]{ocsort}
Jinkun Cao, Jiangmiao Pang, Xinshuo Weng, Rawal Khirodkar, and Kris Kitani.
\newblock Observation-centric sort: Rethinking sort for robust multi-object tracking.
\newblock In \emph{CVPR}, pages 9686--9696, 2023.

\bibitem[Chen et~al.(2015)Chen, Seff, Kornhauser, and Xiao]{driving}
Chenyi Chen, Ari Seff, Alain Kornhauser, and Jianxiong Xiao.
\newblock {DeepDriving}: Learning affordance for direct perception in autonomous driving.
\newblock In \emph{ICCV}, pages 2722--2730, 2015.

\bibitem[Cui et~al.(2023)Cui, Zeng, Zhao, Yang, Wu, and Wang]{sportsmot}
Yutao Cui, Chenkai Zeng, Xiaoyu Zhao, Yichun Yang, Gangshan Wu, and Limin Wang.
\newblock Sportsmot: A large multi-object tracking dataset in multiple sports scenes.
\newblock \emph{arXiv preprint arXiv:2304.05170}, 2023.

\bibitem[Dendorfer et~al.(2020)Dendorfer, Rezatofighi, Milan, Shi, Cremers, Reid, Roth, Schindler, and Leal-Taix{\'e}]{mot20}
Patrick Dendorfer, Hamid Rezatofighi, Anton Milan, Javen Shi, Daniel Cremers, Ian Reid, Stefan Roth, Konrad Schindler, and Laura Leal-Taix{\'e}.
\newblock {MOT20}: A benchmark for multi object tracking in crowded scenes.
\newblock \emph{arXiv preprint arXiv:2003.09003}, 2020.

\bibitem[Dong et~al.(2023{\natexlab{a}})Dong, Cao, and Fu]{dong2023rethinking}
Qiaole Dong, Chenjie Cao, and Yanwei Fu.
\newblock Rethinking optical flow from geometric matching consistent perspective.
\newblock In \emph{CVPR}, pages 1337--1347, 2023{\natexlab{a}}.

\bibitem[Dong et~al.(2023{\natexlab{b}})Dong, Wang, Zhou, and Hua]{dong2023sparse}
Yonghao Dong, Le Wang, Sanping Zhou, and Gang Hua.
\newblock Sparse instance conditioned multimodal trajectory prediction.
\newblock In \emph{ICCV}, pages 9763--9772, 2023{\natexlab{b}}.

\bibitem[Dong et~al.(2024)Dong, Wang, Zhou, Hua, and Sun]{dong2024recurrent}
Yonghao Dong, Le Wang, Sanping Zhou, Gang Hua, and Changyin Sun.
\newblock Recurrent aligned network for generalized pedestrian trajectory prediction.
\newblock \emph{arXiv preprint arXiv:2403.05810}, 2024.

\bibitem[Dosovitskiy et~al.(2015)Dosovitskiy, Fischer, Ilg, Hausser, Hazirbas, Golkov, Van Der~Smagt, Cremers, and Brox]{dosovitskiy2015flownet}
Alexey Dosovitskiy, Philipp Fischer, Eddy Ilg, Philip Hausser, Caner Hazirbas, Vladimir Golkov, Patrick Van Der~Smagt, Daniel Cremers, and Thomas Brox.
\newblock Flownet: Learning optical flow with convolutional networks.
\newblock In \emph{ICCV}, pages 2758--2766, 2015.

\bibitem[Du et~al.(2023)Du, Zhao, Song, Zhao, Su, Gong, and Meng]{strongsort}
Yunhao Du, Zhicheng Zhao, Yang Song, Yanyun Zhao, Fei Su, Tao Gong, and Hongying Meng.
\newblock {StrongSORT}: Make deepsort great again.
\newblock \emph{IEEE T-MM}, 2023.

\bibitem[Ge et~al.(2021)Ge, Liu, Wang, Li, and Sun]{ge2021yolox}
Zheng Ge, Songtao Liu, Feng Wang, Zeming Li, and Jian Sun.
\newblock {YOLOX}: Exceeding yolo series in 2021.
\newblock \emph{arXiv preprint arXiv:2107.08430}, 2021.

\bibitem[Geiger et~al.(2013)Geiger, Lenz, Stiller, and Urtasun]{kitti}
Andreas Geiger, Philip Lenz, Christoph Stiller, and Raquel Urtasun.
\newblock Vision meets robotics: The kitti dataset.
\newblock \emph{IJRR}, 32\penalty0 (11):\penalty0 1231--1237, 2013.

\bibitem[Han et~al.(2022)Han, Huang, Wang, Yu, Liu, and Pan]{Motion-aware}
Shoudong Han, Piao Huang, Hongwei Wang, En Yu, Donghaisheng Liu, and Xiaofeng Pan.
\newblock {MAT}: Motion-aware multi-object tracking.
\newblock \emph{Neurocomputing}, 476:\penalty0 75--86, 2022.

\bibitem[He et~al.(2021)He, Huang, Wang, and Zhang]{he2021learnable}
Jiawei He, Zehao Huang, Naiyan Wang, and Zhaoxiang Zhang.
\newblock Learnable graph matching: Incorporating graph partitioning with deep feature learning for multiple object tracking.
\newblock In \emph{CVPR}, pages 5299--5309, 2021.

\bibitem[He et~al.(2017)He, Gkioxari, Doll{\'a}r, and Girshick]{he2017mask}
Kaiming He, Georgia Gkioxari, Piotr Doll{\'a}r, and Ross Girshick.
\newblock Mask {R-CNN}.
\newblock In \emph{ICCV}, pages 2961--2969, 2017.

\bibitem[Hou et~al.(2021)Hou, Zhang, Sarkis, Bi, Tong, and Liu]{hou2021towards}
Andrew Hou, Ze Zhang, Michel Sarkis, Ning Bi, Yiying Tong, and Xiaoming Liu.
\newblock Towards high fidelity face relighting with realistic shadows.
\newblock In \emph{CVPR}, pages 14719--14728, 2021.

\bibitem[Hui et~al.(2018)Hui, Tang, and Loy]{hui2018liteflownet}
Tak-Wai Hui, Xiaoou Tang, and Chen~Change Loy.
\newblock Liteflownet: A lightweight convolutional neural network for optical flow estimation.
\newblock In \emph{CVPR}, pages 8981--8989, 2018.

\bibitem[Hui et~al.(2020)Hui, Tang, and Loy]{hui2020lightweight}
Tak-Wai Hui, Xiaoou Tang, and Chen~Change Loy.
\newblock A lightweight optical flow cnn—revisiting data fidelity and regularization.
\newblock \emph{IEEE T-PAMI}, 43\penalty0 (8):\penalty0 2555--2569, 2020.

\bibitem[Kalman(1960)]{kalman1960new}
Rudolph~Emil Kalman.
\newblock A new approach to linear filtering and prediction problems.
\newblock 1960.

\bibitem[Khurana et~al.(2021)Khurana, Dave, and Ramanan]{Detecting-invisible-people}
Tarasha Khurana, Achal Dave, and Deva Ramanan.
\newblock Detecting invisible people.
\newblock In \emph{ICCV}, pages 3174--3184, 2021.

\bibitem[Kim et~al.(2021)Kim, Fuxin, Alotaibi, and Rehg]{kim2021discriminative}
Chanho Kim, Li Fuxin, Mazen Alotaibi, and James~M Rehg.
\newblock Discriminative appearance modeling with multi-track pooling for real-time multi-object tracking.
\newblock In \emph{CVPR}, 2021.

\bibitem[Kuhn(1955)]{hungarian}
Harold~W Kuhn.
\newblock The hungarian method for the assignment problem.
\newblock \emph{Naval research logistics quarterly}, pages 83--97, 1955.

\bibitem[Li et~al.(2022)Li, Danelljan, Ding, Huang, and Yu]{li2022tracking}
Siyuan Li, Martin Danelljan, Henghui Ding, Thomas~E Huang, and Fisher Yu.
\newblock Tracking every thing in the wild.
\newblock In \emph{ECCV}, pages 498--515, 2022.

\bibitem[Li et~al.(2023)Li, Zhou, Qin, Wang, Wang, and Zheng]{li2023single}
Yizhe Li, Sanping Zhou, Zheng Qin, Le Wang, Jinjun Wang, and Nanning Zheng.
\newblock Single-shot and multi-shot feature learning for multi-object tracking.
\newblock \emph{arXiv preprint arXiv:2311.10382}, 2023.

\bibitem[Loshchilov and Hutter(2017)]{loshchilov2017decoupled}
Ilya Loshchilov and Frank Hutter.
\newblock Decoupled weight decay regularization.
\newblock \emph{arXiv preprint arXiv:1711.05101}, 2017.

\bibitem[Lowe(2004)]{lowe2004distinctive}
David~G Lowe.
\newblock Distinctive image features from scale-invariant keypoints.
\newblock \emph{IJCV}, 60:\penalty0 91--110, 2004.

\bibitem[Luiten et~al.(2021)Luiten, Osep, Dendorfer, Torr, Geiger, Leal-Taix{\'e}, and Leibe]{luiten2021hota}
Jonathon Luiten, Aljosa Osep, Patrick Dendorfer, Philip Torr, Andreas Geiger, Laura Leal-Taix{\'e}, and Bastian Leibe.
\newblock {HOTA}: A higher order metric for evaluating multi-object tracking.
\newblock \emph{IJCV}, 129:\penalty0 548--578, 2021.

\bibitem[Mancusi et~al.(2023)Mancusi, Panariello, Porrello, Fabbri, Calderara, and Cucchiara]{mancusi2023trackflow}
Gianluca Mancusi, Aniello Panariello, Angelo Porrello, Matteo Fabbri, Simone Calderara, and Rita Cucchiara.
\newblock {DARTHTrackFlow}: Multi-object tracking with normalizing flows.
\newblock In \emph{ICCV}, pages 9531--9543, 2023.

\bibitem[Milan et~al.(2016)Milan, Leal-Taix{\'e}, Reid, Roth, and Schindler]{mot16}
Anton Milan, Laura Leal-Taix{\'e}, Ian Reid, Stefan Roth, and Konrad Schindler.
\newblock {MOT16}: A benchmark for multi-object tracking.
\newblock \emph{arXiv preprint arXiv:1603.00831}, 2016.

\bibitem[Narayanan et~al.(2020)Narayanan, Manoghar, Prashanth~RV, Pham, and Bera]{Robot}
Venkatraman Narayanan, Bala~Murali Manoghar, Rama Prashanth~RV, Phu Pham, and Aniket Bera.
\newblock Seeknet: Improved human instance segmentation and tracking via reinforcement learning based optimized robot relocation.
\newblock \emph{arXiv e-prints}, 2020.

\bibitem[Ng and Henikoff(2003)]{ng2003sift}
Pauline~C Ng and Steven Henikoff.
\newblock {SIFT}: Predicting amino acid changes that affect protein function.
\newblock \emph{Nucleic acids research}, 31\penalty0 (13):\penalty0 3812--3814, 2003.

\bibitem[Oh et~al.(2011)Oh, Hoogs, Perera, Cuntoor, Chen, Lee, Mukherjee, Aggarwal, Lee, Davis, et~al.]{surveil}
Sangmin Oh, Anthony Hoogs, Amitha Perera, Naresh Cuntoor, Chia-Chih Chen, Jong~Taek Lee, Saurajit Mukherjee, JK Aggarwal, Hyungtae Lee, Larry Davis, et~al.
\newblock A large-scale benchmark dataset for event recognition in surveillance video.
\newblock In \emph{CVPR}, pages 3153--3160, 2011.

\bibitem[Pang et~al.(2021{\natexlab{a}})Pang, Qiu, Li, Chen, Li, Darrell, and Yu]{pang2021quasi}
Jiangmiao Pang, Linlu Qiu, Xia Li, Haofeng Chen, Qi Li, Trevor Darrell, and Fisher Yu.
\newblock Quasi-dense similarity learning for multiple object tracking.
\newblock In \emph{CVPR}, pages 164--173, 2021{\natexlab{a}}.

\bibitem[Pang et~al.(2021{\natexlab{b}})Pang, Qiu, Li, Chen, Li, Darrell, and Yu]{quasi}
Jiangmiao Pang, Linlu Qiu, Xia Li, Haofeng Chen, Qi Li, Trevor Darrell, and Fisher Yu.
\newblock Quasi-dense similarity learning for multiple object tracking.
\newblock In \emph{CVPR}, pages 164--173, 2021{\natexlab{b}}.

\bibitem[Qin et~al.(2023)Qin, Zhou, Wang, Duan, Hua, and Tang]{motiontrack}
Zheng Qin, Sanping Zhou, Le Wang, Jinghai Duan, Gang Hua, and Wei Tang.
\newblock {MotionTrack}: Learning robust short-term and long-term motions for multi-object tracking.
\newblock In \emph{CVPR}, pages 17939--17948, 2023.

\bibitem[Ranjan and Black(2017)]{ranjan2017optical}
Anurag Ranjan and Michael~J Black.
\newblock Optical flow estimation using a spatial pyramid network.
\newblock In \emph{CVPR}, pages 4161--4170, 2017.

\bibitem[Ren et~al.(2023)Ren, Han, Ding, Zhang, Wang, and Wang]{focus}
Hao Ren, Shoudong Han, Huilin Ding, Ziwen Zhang, Hongwei Wang, and Faquan Wang.
\newblock Focus on details: Online multi-object tracking with diverse fine-grained representation.
\newblock In \emph{CVPR}, pages 11289--11298, 2023.

\bibitem[Revaud et~al.(2019)Revaud, Weinzaepfel, De~Souza, Pion, Csurka, Cabon, and Humenberger]{revaud2019r2d2}
Jerome Revaud, Philippe Weinzaepfel, C{\'e}sar De~Souza, Noe Pion, Gabriela Csurka, Yohann Cabon, and Martin Humenberger.
\newblock R2d2: repeatable and reliable detector and descriptor.
\newblock \emph{arXiv preprint arXiv:1906.06195}, 2019.

\bibitem[Saleh et~al.(2021)Saleh, Aliakbarian, Rezatofighi, Salzmann, and Gould]{probabilistic}
Fatemeh Saleh, Sadegh Aliakbarian, Hamid Rezatofighi, Mathieu Salzmann, and Stephen Gould.
\newblock Probabilistic tracklet scoring and inpainting for multiple object tracking.
\newblock In \emph{CVPR}, pages 14329--14339, 2021.

\bibitem[Sarlin et~al.(2020)Sarlin, DeTone, Malisiewicz, and Rabinovich]{sarlin2020superglue}
Paul-Edouard Sarlin, Daniel DeTone, Tomasz Malisiewicz, and Andrew Rabinovich.
\newblock Superglue: Learning feature matching with graph neural networks.
\newblock In \emph{CVPR}, pages 4938--4947, 2020.

\bibitem[Segu et~al.(2023)Segu, Schiele, and Yu]{segu2023darth}
Mattia Segu, Bernt Schiele, and Fisher Yu.
\newblock {DARTH}: Holistic test-time adaptation for multiple object tracking.
\newblock In \emph{ICCV}, pages 9717--9727, 2023.

\bibitem[Seidenschwarz et~al.(2023)Seidenschwarz, Bras{\'o}, Serrano, Elezi, and Leal-Taix{\'e}]{seidenschwarz2023simple}
Jenny Seidenschwarz, Guillem Bras{\'o}, V{\'\i}ctor~Castro Serrano, Ismail Elezi, and Laura Leal-Taix{\'e}.
\newblock Simple cues lead to a strong multi-object tracker.
\newblock In \emph{CVPR}, pages 13813--13823, 2023.

\bibitem[Sun et~al.(2018)Sun, Yang, Liu, and Kautz]{sun2018pwc}
Deqing Sun, Xiaodong Yang, Ming-Yu Liu, and Jan Kautz.
\newblock Pwc-net: Cnns for optical flow using pyramid, warping, and cost volume.
\newblock In \emph{CVPR}, pages 8934--8943, 2018.

\bibitem[Sun et~al.(2019)Sun, Yang, Liu, and Kautz]{sun2019models}
Deqing Sun, Xiaodong Yang, Ming-Yu Liu, and Jan Kautz.
\newblock Models matter, so does training: An empirical study of cnns for optical flow estimation.
\newblock \emph{IEEE T-PAMI}, 42\penalty0 (6):\penalty0 1408--1423, 2019.

\bibitem[Sun et~al.(2022)Sun, Cao, Jiang, Yuan, Bai, Kitani, and Luo]{dancetrack}
Peize Sun, Jinkun Cao, Yi Jiang, Zehuan Yuan, Song Bai, Kris Kitani, and Ping Luo.
\newblock {DanceTrack}: Multi-object tracking in uniform appearance and diverse motion.
\newblock In \emph{CVPR}, pages 20993--21002, 2022.

\bibitem[Teed and Deng(2020)]{raft}
Zachary Teed and Jia Deng.
\newblock {RAFT}: Recurrent all-pairs field transforms for optical flow.
\newblock In \emph{ECCV}, pages 402--419, 2020.

\bibitem[Tokmakov et~al.(2021)Tokmakov, Li, Burgard, and Gaidon]{permatrack}
Pavel Tokmakov, Jie Li, Wolfram Burgard, and Adrien Gaidon.
\newblock Learning to track with object permanence.
\newblock In \emph{ICCV}, pages 10840--10849, 2021.

\bibitem[Wang et~al.(2021)Wang, Zheng, Pan, and Xu]{wang2021multiple}
Qiang Wang, Yun Zheng, Pan Pan, and Yinghui Xu.
\newblock Multiple object tracking with correlation learning.
\newblock In \emph{CVPR}, pages 3876--3886, 2021.

\bibitem[Wang et~al.(2023)Wang, Chang, Cai, Li, Hariharan, Holynski, and Snavely]{wang2023omnimotion}
Qianqian Wang, Yen-Yu Chang, Ruojin Cai, Zhengqi Li, Bharath Hariharan, Aleksander Holynski, and Noah Snavely.
\newblock Tracking everything everywhere all at once.
\newblock In \emph{ICCV}, 2023.

\bibitem[Wei et~al.(2020)Wei, Wang, and Huang]{f3net}
Jun Wei, Shuhui Wang, and Qingming Huang.
\newblock {F$^3$Net}: fusion, feedback and focus for salient object detection.
\newblock In \emph{AAAI}, pages 12321--12328, 2020.

\bibitem[Wojke et~al.(2017)Wojke, Bewley, and Paulus]{deepsort}
Nicolai Wojke, Alex Bewley, and Dietrich Paulus.
\newblock Simple online and realtime tracking with a deep association metric.
\newblock In \emph{ICIP}, pages 3645--3649, 2017.

\bibitem[Yan et~al.(2022)Yan, Jiang, Sun, Wang, Yuan, Luo, and Lu]{yan2022towards}
Bin Yan, Yi Jiang, Peize Sun, Dong Wang, Zehuan Yuan, Ping Luo, and Huchuan Lu.
\newblock Towards grand unification of object tracking.
\newblock In \emph{ECCV}, pages 733--751, 2022.

\bibitem[Yu et~al.(2022)Yu, Li, and Han]{yu2022towards}
En Yu, Zhuoling Li, and Shoudong Han.
\newblock Towards discriminative representation: Multi-view trajectory contrastive learning for online multi-object tracking.
\newblock In \emph{CVPR}, pages 8834--8843, 2022.

\bibitem[Yu et~al.(2020)Yu, Chen, Wang, Xian, Chen, Liu, Madhavan, and Darrell]{bdd100k}
Fisher Yu, Haofeng Chen, Xin Wang, Wenqi Xian, Yingying Chen, Fangchen Liu, Vashisht Madhavan, and Trevor Darrell.
\newblock {BDD100K}: A diverse driving dataset for heterogeneous multitask learning.
\newblock In \emph{CVPR}, pages 2633--2642, 2020.

\bibitem[Zeng et~al.(2022)Zeng, Dong, Zhang, Wang, Zhang, and Wei]{zeng2022motr}
Fangao Zeng, Bin Dong, Yuang Zhang, Tiancai Wang, Xiangyu Zhang, and Yichen Wei.
\newblock Motr: End-to-end multiple-object tracking with transformer.
\newblock In \emph{ECCV}, pages 659--675, 2022.

\bibitem[Zhang et~al.(2020)Zhang, Zhou, Chang, Wan, Wang, Wu, and Huang]{fft}
Jimuyang Zhang, Sanping Zhou, Xin Chang, Fangbin Wan, Jinjun Wang, Yang Wu, and Dong Huang.
\newblock Multiple object tracking by flowing and fusing.
\newblock \emph{arXiv preprint arXiv:2001.11180}, 2020.

\bibitem[Zhang et~al.(2021)Zhang, Wang, Wang, Zeng, and Liu]{fairmot}
Yifu Zhang, Chunyu Wang, Xinggang Wang, Wenjun Zeng, and Wenyu Liu.
\newblock {FairMOT}: On the fairness of detection and re-identification in multiple object tracking.
\newblock \emph{IJCV}, 129:\penalty0 3069--3087, 2021.

\bibitem[Zhang et~al.(2022)Zhang, Sun, Jiang, Yu, Weng, Yuan, Luo, Liu, and Wang]{bytetrack}
Yifu Zhang, Peize Sun, Yi Jiang, Dongdong Yu, Fucheng Weng, Zehuan Yuan, Ping Luo, Wenyu Liu, and Xinggang Wang.
\newblock {ByteTrack}: Multi-object tracking by associating every detection box.
\newblock In \emph{ECCV}, pages 1--21, 2022.

\bibitem[Zhang et~al.(2023)Zhang, Wang, and Zhang]{motrv2}
Yuang Zhang, Tiancai Wang, and Xiangyu Zhang.
\newblock Motrv2: Bootstrapping end-to-end multi-object tracking by pretrained object detectors.
\newblock In \emph{CVPR}, pages 22056--22065, 2023.

\bibitem[Zheng et~al.(2023)Zheng, Harley, Shen, Wetzstein, and Guibas]{zheng2023pointodyssey}
Yang Zheng, Adam~W Harley, Bokui Shen, Gordon Wetzstein, and Leonidas~J Guibas.
\newblock {Pointodyssey}: A large-scale synthetic dataset for long-term point tracking.
\newblock In \emph{ICCV}, pages 19855--19865, 2023.

\bibitem[Zhou et~al.(2020)Zhou, Koltun, and Kr{\"a}henb{\"u}hl]{centertrack}
Xingyi Zhou, Vladlen Koltun, and Philipp Kr{\"a}henb{\"u}hl.
\newblock Tracking objects as points.
\newblock In \emph{ECCV}, pages 474--490, 2020.

\bibitem[Zhou et~al.(2022)Zhou, Yin, Koltun, and Kr{\"a}henb{\"u}hl]{GTR}
Xingyi Zhou, Tianwei Yin, Vladlen Koltun, and Philipp Kr{\"a}henb{\"u}hl.
\newblock Global tracking transformers.
\newblock In \emph{CVPR}, pages 8771--8780, 2022.

\end{thebibliography}
}
\clearpage
\setcounter{page}{1}

\section*{Appendix}
We provide more details on tracking scenario attributes in~\cref{sec:Definition of Tracking Scenario Attributes}. The inference speed comparison is conducted in~\cref{sec:Speed}. Architecture and training details are described in~\cref{sec:Architecture and Training Details}. Additionally, a comprehensive version of 'Related Works' and 'Benchmark Evaluation' from the main paper can be found in~\cref{sec:Elaborate Version of TbD in Related Works.} and ~\cref{sec:Detailed Analysis on Benchmarks.}, respectively.

\section{Details of Tracking Scenario Attributes.}
\label{sec:Definition of Tracking Scenario Attributes}
There are countless application scenarios in the world, each presenting unique characteristics. Designing an effective tracker requires the identification of factors with a significant impact on tracking, while disregarding those with minimal influence. To delve into the nature of these scenarios, a more concrete study and analysis are essential. In the following sections, we will provide definitions and quantitative calculations for each attribute.

\subsection{Measurement Metric and Results}
\begin{figure*}[t]
\centering
\includegraphics[width=1\linewidth]{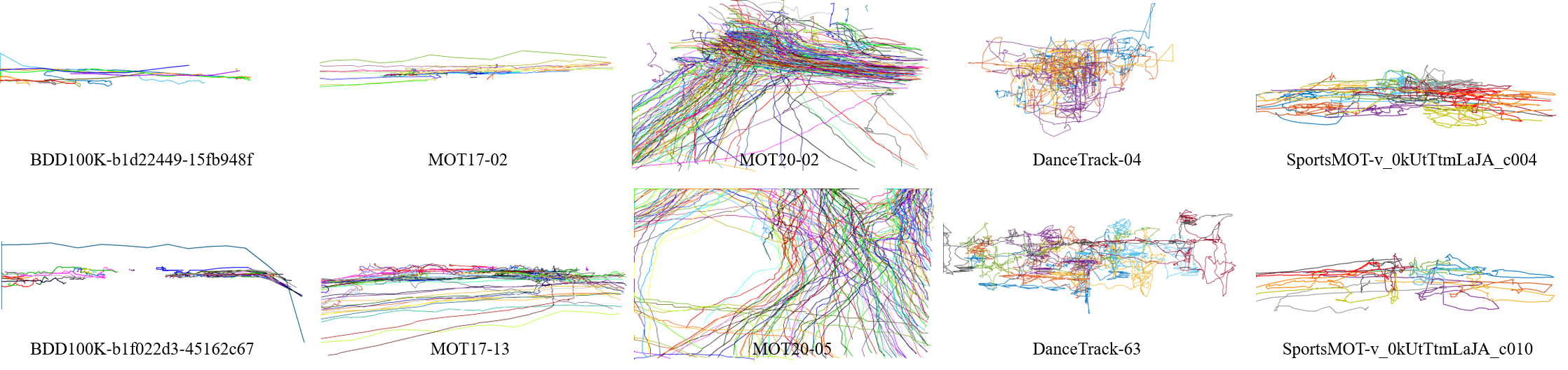}
\caption{Qualitative comparison of motion complexity. Different colors represent the trajectories of different targets. The trajectories in both autonomous driving dataset~(BDD100K) and pedestrian dataset~(MOT17, MOT20) are linear and more predictable than the dancing and sports datasets~(DanceTrack, SportsMOT).
}
\label{fig:MOTION_}
\end{figure*}

\begin{itemize}
	\item \noindent {\bf Motion Complexity.} This metric reflects the irregularity and unpredictability of target motion within the scenario. In our assessments, we decompose motion into direction and velocity. For motion velocity, we calculate the variance of successive velocity magnitudes for each target. For the direction of motion, we transform the continuous direction of the target into the polar coordinate form and calculate the direction mean and variance in the polar coordinate system. The final weighted sum of the two components is the motion complexity.
	\item \noindent {\bf Variation Amplitude.}
This metric reflects the magnitude of the target's variation, which consists of two components: shape variation and absolute position variation. For the former, we obtain the variance of the target's successive aspect ratios. For the latter, we calculate the magnitude of the target's movement relative to its own size. The final weighted sum of the two components is the variation amplitude.
    \item \noindent {\bf Target Density.}
This metric reflects the density of the crowd inside the scene, implicitly reflecting the degree of occlusion between the crowds. For a frame, we calculate the distance between each target in it and measure it by the average body size of the targets. Then we consider people to be occluded by each other when the distance between them is less than half of their body size. More generally, after averaging, this attribute represents the amount of occlusion per capita.
    \item \noindent {\bf Small Target.}
This metric represents the average content of small targets in the dataset. We use the target area to filter small targets with a certain threshold and count the average number of small targets in the scene. 
    \item \noindent {\bf Frame Rate.}
This metric is the number of frames captured in one second of the input video stream. The larger the frame rate, the more information changes within the scene and the more difficult it is to tracking.
\end{itemize}
Based on the definitions above, we measured these attributes on five datasets. The results are shown in~\Cref{table9}. We normalized each attribute as shown in~\Cref{table10}.
We provide a qualitative comparison of motion complexity, displayed in~\Cref{fig:MOTION_}.

\begin{table}[t]
	\centering
	\renewcommand{\arraystretch}{1.0}
	\setlength{\tabcolsep}{2.5pt}
	\Huge
	\resizebox{1\linewidth}{!}{
		\setlength{\tabcolsep}{0.5em}%
		
		\begin{tabular}{lccccc}
			\hline
			\multicolumn{1}{c}{\textbf{\begin{tabular}[c]{@{}c@{}}Scenario\\ Attribute\end{tabular}}} & \textbf{\begin{tabular}[c]{@{}c@{}}Motion\\ Complexity\end{tabular}} & \textbf{\begin{tabular}[c]{@{}c@{}}Variation\\ Amplitude\end{tabular}} & \textbf{\begin{tabular}[c]{@{}c@{}}Target\\ Density\end{tabular}} & \textbf{\begin{tabular}[c]{@{}c@{}}Frame\\ Rate\end{tabular}} & \textbf{\begin{tabular}[c]{@{}c@{}}Small\\ Target\end{tabular}} \\ \hline
			\multicolumn{1}{l|}{BDD100K}                                                              & 1.76                                                                 & 1.80                                                                  & 0.90                                                              & 5                                                             & 7.11                                                            \\
			\multicolumn{1}{l|}{SportsMOT}                                                            & 3.10                                                                 & 0.28                                                                  & 0.48                                                              & 25                                                            & 2.06                                                            \\
			\multicolumn{1}{l|}{MOT17}                                                                & 1.19                                                                 & 0.03                                                                  & 2.77                                                              & 30                                                            & 7.51                                                            \\
			\multicolumn{1}{l|}{MOT20}                                                                & 0.57                                                                 & 0.02                                                                  & 3.30                                                              & 30                                                            & 8.39                                                            \\
			\multicolumn{1}{l|}{DanceTrack}                                                           & 3.44                                                                 & 1.34                                                                  & 1.75                                                              & 30                                                            & 0.00                                                            \\ \hline
		\end{tabular}
		
	}
	\caption{Scores on tracking scenario attributes on five datasets. }  
	\label{table9}  
\end{table}

\begin{table}[t]
	\centering
	\renewcommand{\arraystretch}{1.0}
	\setlength{\tabcolsep}{2.5pt}
	\Huge
	\resizebox{1\linewidth}{!}{
		\setlength{\tabcolsep}{0.5em}%
		
		\begin{tabular}{lccccc}
			\hline
			\multicolumn{1}{c}{\textbf{\begin{tabular}[c]{@{}c@{}}Scenario\\ Attribute\end{tabular}}} & \textbf{\begin{tabular}[c]{@{}c@{}}Motion\\ Complexity\end{tabular}} & \textbf{\begin{tabular}[c]{@{}c@{}}Variation\\ Amplitude\end{tabular}} & \textbf{\begin{tabular}[c]{@{}c@{}}Target\\ Density\end{tabular}} & \textbf{\begin{tabular}[c]{@{}c@{}}Frame\\ Rate\end{tabular}} & \textbf{\begin{tabular}[c]{@{}c@{}}Small\\ Target\end{tabular}} \\ \hline
			\multicolumn{1}{l|}{BDD100K}                                                              & 0.41                                                                 & 1.00                                                                  & 0.15                                                              & 1.00                                                          & 0.85                                                            \\
			\multicolumn{1}{l|}{SportsMOT}                                                            & 0.88                                                                 & 0.06                                                                  & 0.00                                                              & 0.20                                                          & 0.25                                                            \\
			\multicolumn{1}{l|}{MOT17}                                                                & 0.22                                                                 & 0.00                                                                  & 0.81                                                              & 0.00                                                          & 0.90                                                            \\
			\multicolumn{1}{l|}{MOT20}                                                                & 0.00                                                                 & 0.00                                                                  & 1.00                                                              & 0.00                                                          & 1.00                                                            \\
			\multicolumn{1}{l|}{DanceTrack}                                                           & 1.00                                                                 & 0.55                                                                  & 0.45                                                              & 0.00                                                          & 0.00                                                            \\ \hline
		\end{tabular}
		
	}
	\caption{Normalization of detailed scores on tracking scenario attributes on five datasets. }  
	\label{table10}  
\end{table}

\begin{table}[t]
\centering
\renewcommand{\arraystretch}{1.2}
\setlength{\tabcolsep}{2.5pt}
\Huge
\resizebox{1\linewidth}{!}{
    \setlength{\tabcolsep}{0.3em}%
\begin{tabular}{l|ccccc}
\hline
             & \textbf{BDD100K}            & \textbf{SportsMOT} & \textbf{DanceTrack}        & \textbf{MOT17}             & \textbf{MOT20}             \\ \hline
BytrTrack~\cite{bytetrack}    & -                           & 21.1               & 22.4                       & 21.3                       & 15.3                       \\
MOTRV2~\cite{motrv2}       & {\color[HTML]{009901} 11.2} & -                  & {\color[HTML]{009901} 6.8} & {\color[HTML]{009901} 6.4} & {\color[HTML]{009901} 6.5} \\
GHOST~\cite{seidenschwarz2023simple}        & {\color[HTML]{009901} 11.1} & -                  & 2.7                        & 1.2                        & 0.6                        \\ \hline
Ours         & {\color[HTML]{009901} 18.7} & 13.5               & 14.6                       & 15.6                       & 7.6                        \\
Ours~(Accel) & {\color[HTML]{009901} 28.3} & 19.7               & 19.5                       & 18.5                       & 12.2                       \\ \hline
\end{tabular}
	}
	\caption{Comparison of {\bf FPS} on  multiple datasets. 'Accel' represents the accelerated version. Note that the green color represents the inference speed on the basis of the detection result files and the black color represents the speed of the complete tracking process.}  
	\label{table11}  
\end{table}

\subsection{Candidate Attribute.}
\noindent {\bf Appearance similarity.} This attribute is used to describe the similarity of the appearance of the targets within the scene. Overly similar appearances, such as the target wearing the same dance outfit in DanceTrack, the same jersey in SportsMOT, \textit{etc.}, can interfere with the use of the appearance-dominated method by reducing distinguishability between targets.

The reason we did not choose it as the major attribute is that it has a relatively small impact, compared to the damage that other attributes do to motion and appearance. For example, in DanceTrack~\cite{dancetrack}, irregular motion can be fatal to the motion-based method. But even if the appearance is similar, we can still rely on the appearance-based method to track targets well.

\section{Inference Speed.}
\label{sec:Speed}
As shown in~\Cref{table11}, we give the inference speeds of our GeneralTrack and several commonly used SOTAs on multiple datasets. Note that all results are tested under 1 NVIDIA GeForce RTX 3090 Ti GPU. 
Among all prior trackers, BytrTrack is mostly the fastest, and our accelerated version can achieve inference speeds similar to it, but our performance is better on almost all datasets. 

\noindent {\bf Accelerated Version.} 
Because our Feature Relation Extractor constructs global relationships, reducing the image size can accelerate the inference speed with little impact on performance. So we can resize the input image to a smaller size if there is a need for acceleration.

\section{Architecture and Training Details.}
\label{sec:Architecture and Training Details}
The weights-sharing convolutional neural network in Feature Relation Extractor consists of 6 residual blocks~(2 at 1/2 resolution, 2 at 1/4 resolution, and 2 at 1/8 resolution). For training, we employ the AdamW optimizer~\cite{loshchilov2017decoupled} and limit the gradients to the interval [-1, 1]. The main purpose of dense flow and correspondence tasks is to construct a pixel-wise dense relationship of an image pair, and our task is to construct the pixel-wise relationship and transform them into instance-wise associations from fine to coarse. Therefore, we use KITTI~\cite{kitti} for pre-training in the settings of optical flow for enhancing the capability of Feature Relation Extractor~(the weights-sharing convolutional neural network). 

\noindent {\bf Background Mask.} 
To focus more on foreground targets, we mask the background area to reduce its disturbance
in the calculation of the correlation pyramid. Based on the tracklet bank $\mathbb{T}$ and prior detection $\mathcal{D}^{t}$, respectively, we generate binarized matrices, where each binarized element represents whether it belongs to a foreground target or not. Background mask is only used on DanceTrack in the benchmark results.

\section{Elaborate Version of TbD in Related Works.}
\label{sec:Elaborate Version of TbD in Related Works.}
\noindent{\bf Tracking-by-Detection.}
The dominant paradigm in the field has long been tracking-by-detection~\cite{bytetrack, sort, strongsort, ocsort, motiontrack, fft, he2021learnable, quasi, permatrack, deepsort}, which divides tracking into two steps: (i) frame-wise object detection, (ii) data association to link the detections and form trajectories. The core of the data association is to construct inter-frame relation~(Affinity matrix) between tracklets and detections, and then complete the matching with the Hungarian algorithm~\cite{hungarian}. The affinity matrix for matching is often driven by motion information~\cite{motiontrack, probabilistic, sort, Motion-aware} or appearance information~\cite{deepsort, strongsort, focus, yu2022towards}.
Motion-based trackers exploit the fact that object displacements tend to be small given two neighboring frames. This allows them to leverage spatial proximity for matching with tools such as Kalman filters~\cite{kalman1960new, sort} or its variant version~\cite{fairmot,Motion-aware,Detecting-invisible-people}. Some recent works~\cite{probabilistic, motiontrack} use data-driven motion models for more accurate motion prediction and lead to more robust tracking. Motion trackers work well for pedestrian tracking scenarios with regular motion, \emph{i.e.}, MOT17 and MOT20. But when the frame rate goes low, the movement amplitude gets larger, and the motion becomes more complex, motion wears out, and it's needed for appearance.
Appearance relies on extracting discriminative features to construct instance-level relations. DeepSORT~\cite{deepsort} firstly adopts a stand-alone Re-ID model to extract appearance features from the detection boxes. Follow-up efforts~\cite{kim2021discriminative, wang2021multiple, pang2021quasi, yu2022towards, seidenschwarz2023simple, focus, dong2024recurrent} use a variety of approaches to come up with better appearance models, such as domain adaptation~\cite{seidenschwarz2023simple}, contrastive learning~\cite{quasi}, \emph{etc.}. These appearances rely on distinguishable overall voxel information and can handle the above occasions where motions cannot resolve. However, it has limitations when it encounters occlusion caused by dense crowds or the targets are too small to extract effective features.

In facing these issues, previous methods worked towards a better balance between motion and appearance. Some works~\cite{bytetrack, seidenschwarz2023simple, strongsort, deepsort, botsort} choose whether to emphasize motion or appearance more based on a very strong prior and multiple experiment attempts. TrackFlow~\cite{mancusi2023trackflow} addresses these issues by building on an elegant probabilistic formulation that requires additional virtual datasets for training.
In contrast, we propose a new tracking method that achieves generalization while avoiding the need to balance motion and appearance.

\section{Detailed Analysis on Benchmarks.}
\label{sec:Detailed Analysis on Benchmarks.}
\noindent{\bf BDD100K.}
Our GeneralTrack outperforms the state-of-the-art methods in most key metrics, \emph{i.e.}, ranks first for metrics mTETA, mHOTA, mIDF1, mMOTA, HOTA, IDF1 and ranks second for MOTA in the validation set. On the test set, GeneralTrack achieves the best performance under most of the key metrics, with the rest of the metrics ranked second and very close to the best. Note that we use the same detection results as GHOST and ByteTrack, compared to which brings a big boost~(+1.2 mHOTA, +1.4 HOTA, +0.6 mIDF1, + 1.8 IDF1, +1.5 mMOTA, +0.7 MOTA) in the validation set and (+1.1 mHOTA, +1.5 HOTA, + 1.6 IDF1) in the test set on GHOST; (+1.6 mHOTA, +1.8 HOTA, +1.4 mIDF1, +2.3 IDF1) in the validation set and (+1.1 IDF1, +2.3 IDF1) in the test set on ByteTrack. While Bytetrack selects appearance and GHOST weight summation of motion and appearance, in comparison, our approach outperforms such hand-designed algorithms by a large margin, demonstrating the generalizability of our approach to the multi-class tracking task with a low frame rate.

\noindent{\bf SportsMOT.}
GeneralTrack ranks first in all key metrics~(HOTA, MOTA, IDF1). While using the same detections, we gain significant improvement~(+8.4 HOTA, + 2.3 IDF1) on MixSort-Byte and (+ 2.0 IDF1) on MixSort-OC. Otherwise, MixSort-Byte and MixSort-OC train the association component on both the training set and the validation set; in contrast, we train only on the training set. Even so, we are still surpassing them and the improvements in metrics prove the superiority of our association capabilities even under very severe motion complexity.

\noindent{\bf DanceTrack.}
When being generalized to the dancing dataset, our method outperforms all CNN-based trackers. Note that all these CNN-based methods share the same detection, our GeneralTrack ranks first in HOTA and is 2.3 higher than the second place. Similarly, our AssA and DetA are 2.4 and 0.9 higher than the second place, respectively.
Although our method is inferior to MOTRv2, it uses both YOLOX and MOTR with more than two hundred times training resource usage than ours. And our method outperforms it on several other datasets.
These results indicate that our method is robust to large variation amplitudes of the target in addition to handling complex motions.

{\tiny }\noindent{\bf MOT17 and MOT20.}
Both datasets are pedestrian tracking datasets with regular motion patterns and both have dense crowd distributions and smaller targets. Our method ranks first in all key metrics HOTA, MOTA, IDF1 and DetA, IDs on MOT17 and ranks second in HOTA and MOTA on MOT20. On MOT17, GeneralTrack improves over the best previous methods, \emph{e.g.}, gaining the improvement~(+1.2 HOTA, +1.9 MOTA, + 1.2 IDF1) on GHOST and (+0.9 HOTA, +0.3 MOTA, + 1.0 IDF1) on ByteTrack. For MOT20, our  method performs more stably under three key metrics compared to other trackers.
These results show that our method can generalize well to scenarios where crowded and small targets exist.

\section{More Discussion on Domain Generalization.}
On domain generalization experiments for cross-class on BDD100K,  domain generalization of GHOST~\cite{seidenschwarz2023simple} focuses on the ReID model while our method addresses the association model. We provide the performance of both training with one class and then tracking in the entire dataset (GeneralTrack:mHOTA 46.5, mIDF1 55.2, mMOTA 45.5 ; GHOST: mHOTA 45.7, mIDF1 55.6, mMOTA 44.9). It is worth noting that we only train on one class (car) on BDD100K, whereas GHOST train one class (people) with data outside of BDD100K.

\section{\bf Resolution and Downsampling Scale.} 

\begin{table}[t]
\centering
\renewcommand{\arraystretch}{1.2}
\setlength{\tabcolsep}{6pt}
\Huge
\resizebox{0.8\linewidth}{!}{
\begin{tabular}{cc|ccccccc}
\hline
Inuput 
 Size& DS& mHOTA & mIDF1 & mMOTA & HOTA & IDF1 & MOTA & IDs   \\ \hline
720×1280    & 2                                                          & 47.1  & 56.1  & 46.1  & 63.4 & 72.5 & 68.3 & 8503  \\
720×1280    & 4                                                          & 46.6  & 55.3  & 45.3  & 62.6 & 71.5 & 67.7 & 10283 \\
360×640     & 2                                                          & 46.2  & 54.8  & 44.8  & 62.3 & 71.0 & 67.4 & 9781  \\ \hline
\end{tabular}}
\end{table}

The ablation study below shows that our method is insensitive to both the frame resolution and the downsampling scale  (DS).


\end{document}